\documentclass[sigconf, nonacm]{acmart}

\newcommand\vldbdoi{XX.XX/XXX.XX}
\newcommand\vldbpages{XXX-XXX}
\newcommand\vldbvolume{15}
\newcommand\vldbissue{1}
\newcommand\vldbyear{2022}
\newcommand\vldbauthors{\authors}
\newcommand\vldbtitle{\shorttitle} 
\newcommand\vldbavailabilityurl{URL_TO_YOUR_ARTIFACTS}
\newcommand\vldbpagestyle{plain} 

\PassOptionsToPackage{protrusion, expansion}{microtype} 

\usepackage{amsmath, amsthm}
    
\usepackage{graphicx, xcolor} 
\usepackage{hyperref, url}
\usepackage{adjustbox}
\usepackage{booktabs, multirow}
\usepackage{caption}
\usepackage{subcaption}
\usepackage{enumitem}
\usepackage{lipsum}
\usepackage{setspace}
\usepackage{algorithm, algpseudocode}
    \definecolor{BLUE}{rgb}{.0, .2, .6}
    \definecolor{BLUEalt}{HTML}{1e50a2}
    \definecolor{RED}{HTML}{c9171e}
    \algrenewcommand{\alglinenumber}[1]{{\scriptsize\bfseries\ttfamily\color{RED}#1}}

\usepackage{footnote}
\makesavenoteenv{tabular}
\makesavenoteenv{table}

\setcopyright{none}
\renewcommand\footnotetextcopyrightpermission[1]{}
\usepackage{xspace}
\newcommand{\techname}{COMET\xspace}

\usepackage[zerostyle=d]{newtxtt}

\usepackage[normalem]{ulem}

\begin{document}

\title{\techname: A Novel Memory-Efficient Deep Learning Training Framework by Using Error-Bounded Lossy Compression}


\newcommand{\AFFIL}[4]{%
     \affiliation{%
         \institution{\small #1}
         \city{#2}\state{#3}\country{#4}
     }
     }

\author{Sian Jin}{\AFFIL{Washington State University}{Pullman}{WA}{USA}}
\email{sian.jin@wsu.edu}

\author{Chengming Zhang}{\AFFIL{Washington State University}{Pullman}{WA}{USA}}
\email{chengming.zhang@wsu.edu}


\author{Xintong Jiang}{\AFFIL{McGill University}{Montréal}{QC}{Canada}}
\email{xintong.jiang@mail.mcgill.ca}

\author{Yunhe Feng}{\AFFIL{University of Washington}{Seattle}{WA}{USA}}
\email{yunhe@uw.edu}

\author{Hui Guan}{\AFFIL{University of Massachusetts}{Amherst}{MA}{USA}}
\email{huiguan@cs.umass.edu}

\author{Guanpeng Li}{\AFFIL{University of Iowa}{Iowa City}{IA}{USA}}
\email{guanpeng-li@uiowa.edu}

\author{Shuaiwen Leon Song}{\AFFIL{University of Sydney}{Sydney}{NSW}{Australia}}
\email{shuaiwen.song@sydney.edu.au}

\author{Dingwen Tao}{\AFFIL{Washington State University}{Pullman}{WA}{USA}}
\email{dingwen.tao@wsu.edu}

\begin{abstract}

Deep neural networks (DNNs) are becoming increasingly deeper, wider, and non-linear due to the growing demands on prediction accuracy and analysis quality. Training wide and deep neural networks require large amounts of storage resources such as memory because the intermediate activation data must be saved in the memory during forward propagation and then restored for backward propagation. However, state-of-the-art accelerators such as GPUs are only equipped with very limited memory capacities due to hardware design constraints, which significantly limits the maximum batch size and hence performance speedup when training large-scale DNNs. Traditional memory saving techniques either suffer from \textcolor{black}{performance overhead} or are constrained by limited interconnect bandwidth or specific interconnect technology. 

In this paper, we propose a novel memory-efficient CNN training framework (called \techname) that leverages error-bounded lossy compression to significantly reduce the memory requirement for training in order to allow training larger models or to accelerate training. Different from the state-of-the-art solutions that adopt image-based lossy compressors (such as JPEG) to compress the activation data, our framework purposely adopts error-bounded lossy compression with a strict error-controlling mechanism. Specifically, we perform a theoretical analysis on the compression error propagation from the altered activation data to the gradients, and empirically investigate the impact of altered gradients over the training process. Based on these analyses, we optimize the error-bounded lossy compression and propose an adaptive error-bound control scheme for activation data compression. We evaluate our design against state-of-the-art solutions with five widely-adopted CNNs and ImageNet dataset. Experiments demonstrate that our proposed framework can significantly reduce the training memory consumption by up to 13.5$\times$ over the baseline training and 1.8$\times$ over another state-of-the-art compression-based framework, respectively, with little or no accuracy loss. 


\end{abstract}

\maketitle

\pagestyle{\vldbpagestyle}
\begingroup\small\noindent\raggedright\textbf{PVLDB Reference Format:}\\
\vldbauthors. \vldbtitle. PVLDB, \vldbvolume(\vldbissue): \vldbpages, \vldbyear.\\
\href{https://doi.org/\vldbdoi}{doi:\vldbdoi}
\endgroup
\begingroup
\renewcommand\thefootnote{}\footnote{\noindent
This work is licensed under the Creative Commons BY-NC-ND 4.0 International License. Visit \url{https://creativecommons.org/licenses/by-nc-nd/4.0/} to view a copy of this license. For any use beyond those covered by this license, obtain permission by emailing \href{mailto:info@vldb.org}{info@vldb.org}. Copyright is held by the owner/author(s). Publication rights licensed to the VLDB Endowment. \\
\raggedright Proceedings of the VLDB Endowment, Vol. \vldbvolume, No. \vldbissue\ %
ISSN 2150-8097. \\
\href{https://doi.org/\vldbdoi}{doi:\vldbdoi} \\
}\addtocounter{footnote}{-1}\endgroup

\ifdefempty{\vldbavailabilityurl}{}{
\vspace{.3cm}
\begingroup\small\noindent\raggedright\textbf{PVLDB Artifact Availability:}\\
The source code, data, and/or other artifacts have been made available at \url{https://github.com/jinsian/COMET}.
\endgroup
}


 \setlength{\textfloatsep}{6pt}

\section{Introduction}
\label{sec:introduction}

Deep neural networks (DNNs) have rapidly evolved to the state-of-the-art technique for many artificial intelligence (AI) tasks in various science and technology domains, including image and vision recognition \cite{simonyan2014very}, recommendation systems~\cite{wang2015collaborative}, and natural language processing (NLP)~\cite{collobert2008unified}.
DNNs contain millions of parameters in an unparalleled representation, which is efficient for modeling complex nonlinearities. 
Many works \cite{krizhevsky2012imagenet,szegedy2015going,he2016deep} have suggested that using either deeper or wider DNNs is an effective way to improve analysis quality and in fact, many recent DNNs have gone significantly deeper and/or wider \cite{wang2018superneurons,jin2019deepsz}. 
Most of such wide and deep neural networks contain a large portion of convolutional layers, also known as convolutional neural networks (CNNs).
For instance, EfficientNet-B7 increases the number of convolutional layers from 31 to 109 and doubles the layer width compared to the base EfficientNet-B0 for higher accuracy
(i.e., top-1 accuracy of 84.3\% compared to 77.1\% on the ImageNet dataset) \cite{tan2019efficientnet}.

In this paper, we explore a \textit{general memory-driven approach for enabling efficient deep learning training}. Specifically, our goal is to drastically reduce the memory requirement for training in order to enlarge the limit of maximum batch size for training speedup. 
When training a DNN model, the intermediate activation data (i.e., the input of all the neurons) is typically saved in the memory during forward propagation, and then restored during backpropagation to calculate gradients and update weights accordingly \cite{hecht1992theory}. However, taking into account the deep and wide layers in the current large-scale nonlinear DNNs, storing these activation data from all the layers requires large memory spaces which are not available in state-of-the-art training accelerators such as GPUs.
For instance, in recent climate research \cite{kurth2018exascale}, training DeepLabv3+ neural network with 32 images per batch requires about 170 GB memory, which is about 2$\times$ as large as the memory capacity supported by the latest NVIDIA GPU A100.
Furthermore, modern DNN model design trades off memory requirement for higher accuracy. For example, Gpipe \cite{huang2019gpipe} increases the memory requirement by more than 4$\times$ for achieving a top-1 accuracy improvement of 5\% from Inception-V4.

Evolving in recent years, on the one hand, model-parallel \cite{ben2019demystifying} techniques that distribute the model into multiple nodes can reduce the memory consumption of each node but introduce high communication overheads; on the other hand, data-parallel techniques \cite{sergeev2018horovod} replicate the model in every node but distribute the training data to different nodes, thereby suffering from high memory consumption to fully utilize the computational power. 
Several techniques such as recomputation, migration, and lossless compression of activation data have been proposed to address the memory consumption challenge for training large-to-large-scale DNNs. For example, GeePS~\cite{cui2016geeps} and vDNN~\cite{rhu2016vdnn} have developed data migration techniques for transferring the intermediate data from GPU to CPU to alleviate the memory burden. However, the performance of data migration approaches is limited by the specific intra-node interconnect technology (e.g., PCIe and NVLinks \cite{foley2017ultra}) and its available bandwidth. 
Some other approaches are proposed to recompute the activation data \cite{chen2016training,gomez2017reversible}, but they often incur large performance degradation, especially for computationally intensive layers such as convolutional layers. 
Moreover, memory compression approaches based on lossless compression of activation data \cite{son2014data} suffer from the limited compression ratio (e.g., only around 2:1 for most floating-point data). 
Alternatively, recent works \cite{evans2020jpeg,choukse2020buddy} proposed to develop compression offloading accelerators for reducing the activation data size before transferring it to the CPU DRAM.  
However, adding a new dedicated hardware component to the existing GPU architecture requires tremendous industry efforts and is not ready for immediate deployment. This design may not be general enough to accommodate future DNN models and accelerator architectures.

To tackle these challenges, we propose a memory-efficient deep neural network training framework (called \techname, lossy \underline{C}ompres-sion \underline{O}ptimized \underline{M}emory-\underline{E}fficient \underline{T}raining) by compressing the activation data using adaptive error-bounded lossy compression. 
Compared to lossy compression approaches such as JPEG \cite{wallace1992jpeg} and JPEG2000 \cite{taubman2012jpeg2000}, error-bounded lossy compression can provide more strict control over the errors that occurred to the floating-point activation data. Also, compared to lossless compression such as GZIP \cite{gzip} and Zstd \cite{zstd}, it can offer a much higher compression ratio to gain higher memory consumption reduction and performance improvement. 
The key insights explored in this work include:
(i) the impact of compression errors that occurred in the activation data on the gradients and the entire CNN training process under the strict error-controlling lossy compression can be theoretically and experimentally analyzed, and
(ii) the validation accuracy can be well maintained based on an adaptive fine-grained control over error-bounded lossy compression (i.e., compression error). 
To the best of our knowledge, this is the first work \textit{to investigate the lossy compression error impact during CNN training and leverage this analysis to significantly reduce the memory consumption for training large CNNs while maintaining high validation accuracy}.
In summary, this paper makes the following contributions:
\begin{itemize}[noitemsep, topsep=8pt, leftmargin=1.3em]
    \item We propose a novel memory-efficient CNN training framework via dynamically compressing the intermediate activation data through error-bounded lossy compression.
    \item We provide a thorough analysis of the impact of compression error propagation during DNN training from both theoretical and empirical perspectives.
    \item We propose an adaptive scheme to adaptively configure the error-bounded lossy compression based on a series of current training status data.
    \item We propose an improved SZ error-bounded lossy compression to further handle compressing continuous zeros in the intermediate activation data, which can avoid the significant alteration (vanish or explosion) of gradients.
    \item We evaluate our proposed training framework on four widely-adopted DNN models (AlexNet, VGG-16, ResNet-18, ResNet-50) with the ImageNet-2012 dataset and compare it against state-of-the-art solutions. Experimental results show that our design can reduce the memory consumption by up to 13.5$\times$ and 1.8$\times$ compared to the original training framework and the state-of-the-art method, respectively, under the same batch size. \techname can improve the end-to-end training performance by leveraging the saved memory for some models (e.g., about 2$\times$ training performance improvement on AlexNet).
\end{itemize}
 
 The rest of this paper is organized as follows. In Section~\ref{sec:background}, we discuss the background and motivation of our research. In Section~\ref{sec:design}, we describe an overview of our proposed \techname framework. In Section~\ref{sec:analysis}, we present our theoretical support of error impact on validation accuracy from compressed activation data during training. In Section~\ref{sec:evaluation}, we present the evaluation results of our proposed \techname from the perspectives of parameter selection, memory reduction ability, and performance. In Section~\ref{sec:conclusion}, we conclude our work and discuss our future work.
\section{Background and Motivation}
\label{sec:background}

In this section, we first present the background information on large-scale DNN training (i.e., some related work on memory reduction techniques for training) and error-bounded lossy compression for floating-point data. We then discuss the motivation of this work and our research challenges.

\subsection{Training Large-Scale DNNs}


\begin{figure}[]
    \centering
    \includegraphics[width=1.0\linewidth]{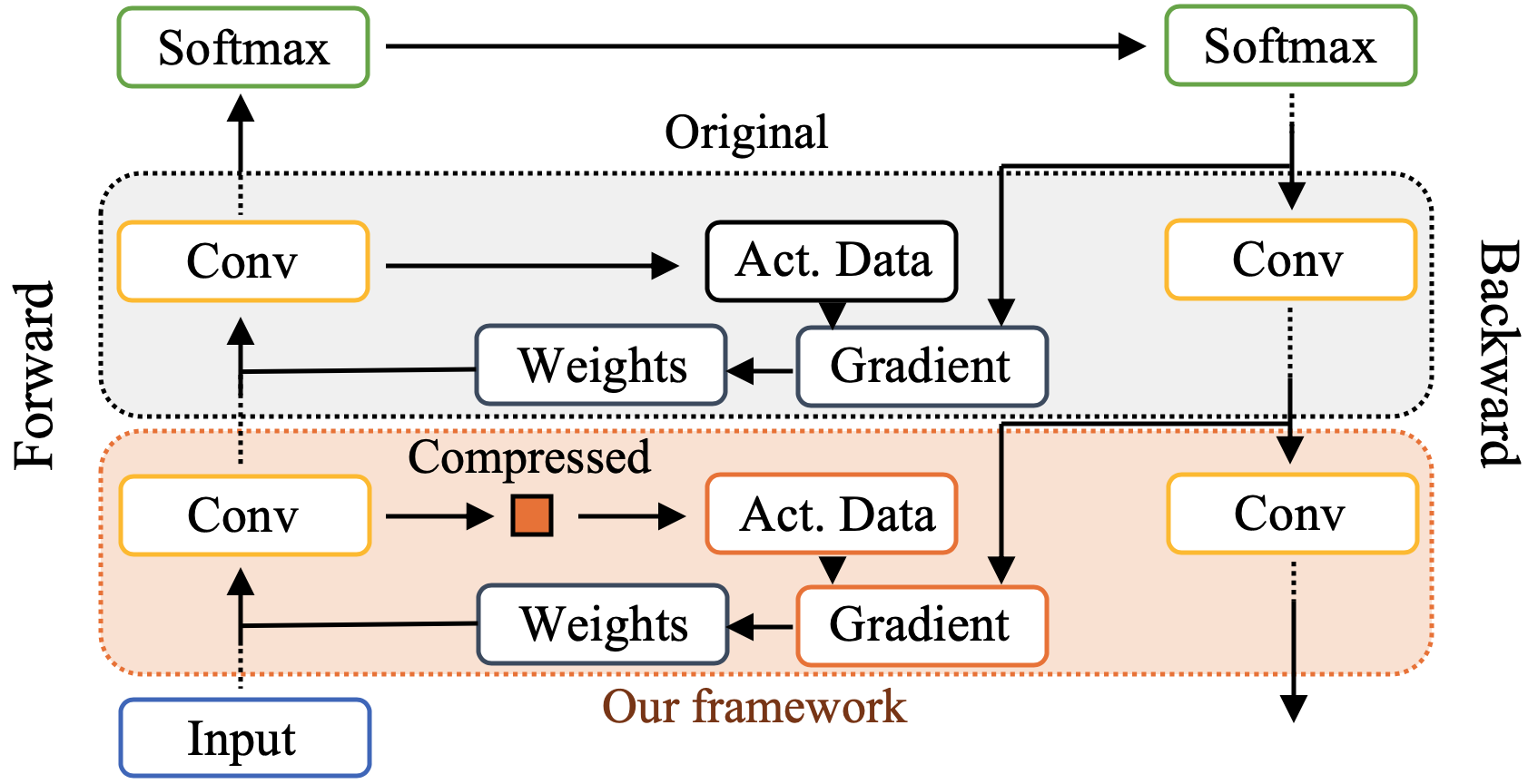}
    \vspace{-4mm}
    \caption{An example data flow of one iteration in CNN training with our \techname framework.}
    \label{fig:fig-1}
\end{figure}

Training deep and wide neural networks has become increasingly challenging.
While many state-of-the-art deep learning frameworks such as TensorFlow \cite{abadi2016tensorflow} and PyTorch \cite{paszke2019pytorch} can provide high training throughput by leveraging the massive parallelism on general-purpose accelerators such as GPUs, one of the most common bottlenecks remains to be the high memory consumption during the training process, especially considering the limited on-chip memory available on modern DNN accelerators. This is mainly due to the ever-increasing size of the activation data computed in the training process. 
Training a neural network involves many training epochs to update and learn the model weights. Each iteration includes a forward and backward propagation, as shown in Figure~\ref{fig:fig-1}. The intermediate activation data (the output from each neuron) generated by every layer are commonly kept in the memory until the backpropagation reaches this layer again. 
Several works \cite{chen2016training, gomez2017reversible,rhu2016vdnn, cui2016geeps,evans2020jpeg} have pointed out the \textcolor{black}{large} gap between the time when the activation data is generated in the forward propagation and the time when the activation data is reused in the backpropagation, especially when training very deep neural networks.

\begin{figure}[]
    \centering
    \includegraphics[width=0.9\linewidth]{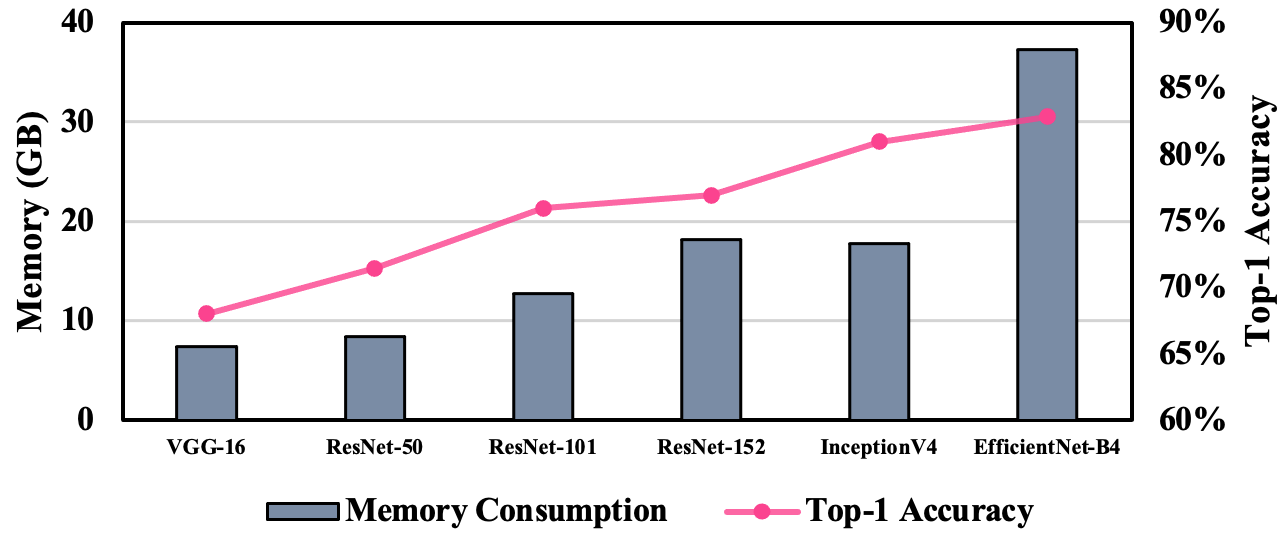}
    \vspace{-3mm}
    \caption{Memory consumption and top-1 accuracy of different state-of-the-art neural networks (batch size = 128).}
    \label{fig:fig-2}
\end{figure}

Figure~\ref{fig:fig-2} shows the memory consumption of various neural networks. For these CNNs, comparing with the model/weight size, the size of activation data is much larger since the convolution kernels are relatively small compared to the activation tensors. 
\textcolor{black}{In addition, training models with enormous batch size over multiple nodes can significantly reduce the training time~\cite{goyal2017accurate, you2019large, pauloski2020convolutional} while reducing the memory consumption can enlarge the maximum batch size capability of a single node for an overall lower cost.}
In summary, we are facing two main challenges due to the high memory consumption in today's deep learning training: 
(1) it is challenging to scale up the training process under a limited GPU memory capacity, and 
(2) a limited batch size leads to low training performance.

In recent years, several works have been proposed to reduce the memory consumption for DNN training, including activation data recomputation \cite{chen2016training, gomez2017reversible}, migration \cite{rhu2016vdnn, cui2016geeps}, and compression \cite{evans2020jpeg}.
Recomputation takes advantage of the layers with low computational cost, such as the pooling layer. 
Specifically, it \textcolor{black}{frees} those layers' activation data and recomputes them based on their prior layers during the backpropagation on demand. 
This method can reduce unnecessary memory costs, but it is only applicable for the layers of limited types to achieve low performance overhead.
For example, compute-intensive convolutional layers that are often hard to recomputed dwarf the efforts of such a method.

Another type of methods are proposed around data migration \cite{rhu2016vdnn, cui2016geeps}, which sends the activation data from the accelerator to the CPU host when generated, and then loads it back from the host when needed. 
However, the performance of data migration heavily depends on the interconnect bandwidth available between the host and the accelerator(s), and the intra-node interconnect technology applied. For example, NVLink \cite{foley2017ultra} technology is currently limited to high-end NVIDIA AI nodes (e.g., DGX series) and IBM power series. 
This paper targets to develop a general technique that can be applied to all types of HPC and datacenter systems. 

\textcolor{black}{Finally}, data compression is another efficient approach to reduce memory consumption, especially for conserving the memory bandwidth \cite{evans2020jpeg, lal20172mc, ekman2005robust}. The basic idea using data compression here is to compress the activation data when generated, hold the compressed data in the memory, and decompress it when needed. 
However, using lossless compression \cite{choukse2020buddy} can only provide a relatively low memory reduction ratio (i.e., compression ratio), e.g., typically lower than 2$\times$. 
Some other studies such as JPEG-ACT \cite{evans2020jpeg} leverages the similarity between activation tensors and images for vision recognition tasks and apply a modified JPEG lossy compressor to activation data. 
But it suffers from two main drawbacks:
First, it introduces uncontrollable compression errors to activation data. Eventually, it could lose control of the overall training accuracy since JPEG is mainly designed for images and is an integer-based lossy compression. 
Second, the JPEG-based solution \cite{evans2020jpeg} needs support from a dedicated hardware component to be added to GPU hardware,
and it cannot be directly deployed to today's systems.

We note that all three methods above are orthogonal to each other, which means they could be deployed together to maximize the \textcolor{black}{memory reduction} and training performance. Thus, in this paper, we mainly focus on designing an efficient data compression, more specifically a lossy-compression-based solution, to achieve the memory reduction ratio beyond the state-of-the-art approach on CNN models. 
In addition, since convolutional layers are the most difficult type of layers for efficient recomputation, our solution focuses on convolutional layers to provide high compression ratios with minimum performance overheads and accuracy losses.  
\textcolor{black}{
We also note that \techname can further improve the training performance by combining with model parallelism techniques such as Cerebro~\cite{nakandala2020cerebro}, which 
is designed for efficiently training multiple model configurations to select the best model configuration. 
}

\subsection{Lossy Compression for Floating-Point Data}
Floating-point data compression has been studied for decades. 
Lossy compression
can compress data with little information loss in the reconstructed data.
Compared to lossless compression, lossy compression can provide a much higher compression ratio while still maintaining useful information for scientific or visualized discoveries. 
Lossy compressors offer different compression modes to control compression error or compression ratio, such as error-bounded mode. 
The error-bounded mode requires users to set an error bound, such as absolute error bound and point-wise relative error bound. 
The compressor ensures that the differences between the original and reconstructed data do not exceed the user-set error bound.

In recent years, a new generation of lossy compressors for scientific data have been proposed and developed, such as SZ~\cite{di2016fast, tao2017significantly, liangerror} and ZFP~\cite{zfp}.
Unlike traditional lossy compressors such as JPEG \cite{wallace1992jpeg} which are designed for images (in integers), SZ and ZFP are designed to compress floating-point data and can provide a strict error-controlling scheme based on user's requirements.
In this work, we choose SZ instead of ZFP because the GPU version of SZ---cuSZ~\cite{tian2020cusz}
\footnote{Compared to CPU SZ, cuSZ can provide much higher compression and decompression speed on GPUs and can also be tuned to avoid the CPU-GPU data transfer overheads.}
---provides a higher compression ratio than ZFP and offers the absolute error bound mode that the GPU version of ZFP does not support (but necessary for our error control).
Specifically, SZ is a prediction-based error-bounded lossy compressor for scientific data. SZ has three main steps: (1) predict each data point's value based on its neighboring points by using an adaptive, best-fit prediction method; (2) quantize the difference between the real value and predicted value based on the user-set error bound; and (3) apply a customized Huffman coding and lossless compression to achieve a higher ratio.
We note that a recent work~\cite{jin2019deepsz} proposed to use the new generation of lossy compressors to compress DNN weights, thereby significantly reducing model storage overhead and transmission time. However, this work only focuses on compressing the DNN model itself instead of compressing the activation data to reduce memory consumption.


\subsection{Research Goals and Challenges}
\label{subsec:challenge}

\begin{figure*}[]
    \centering
    \includegraphics[width=0.9\linewidth]{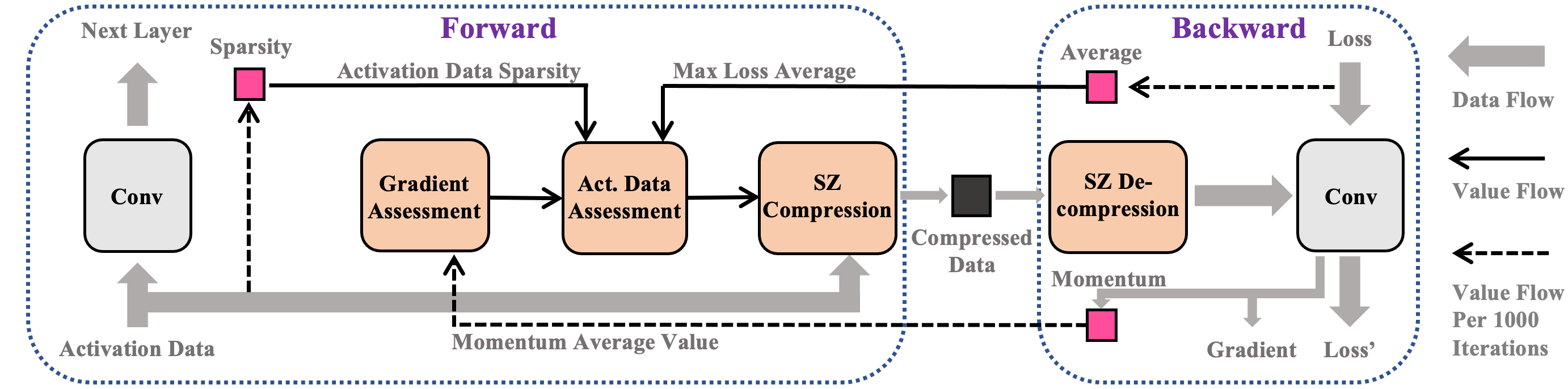}
    \vspace{-2mm}
    \caption{Overview of our proposed memory-efficient CNN training framework \techname.}
    \vspace{-2mm}
    \label{fig:fig-41-1}
\end{figure*}

This is the first known work that explores whether the new generation of lossy compression techniques, which have been widely adopted to help scientific applications gain significant compression ratio with precise error control, can significantly reduce the high memory consumption of the common gradient-descent-based training scenarios (e.g., CNNs with forward and backward propagation). 
We focus on compressing the activation data of convolutional layers, because (1) convolutional layers dominate the size of activation data and cannot be easily recomputed \cite{wang2018superneurons}, and (2) non-convolutional layers (e.g., pool layers or fully-connected layers) can be easily recomputed for memory reduction due to their low computation complexity.

Note that convolutional layers also dominate the computational time during training process, which benefits us for apply compression with low overhead.
Similar to many previous studies \cite{liu2018layrub, evans2020jpeg, wang2018superneurons, rhu2016vdnn}, our research goal is to develop an efficient and generic strategy to achieve a high reduction in memory consumption for CNN training. Our work can increase the batch size limit and convergence speed or enable training on the hardware with lower memory capacity for the same CNN model.

To achieve this goal, there are several critical challenges to be addressed. 
First, \textcolor{black}{because} of the prediction-based mechanism of SZ lossy compressor, when compressing continuous zeros in between the data, SZ cannot guarantee the decompressed array to remain the same continuous zeros but decompress them into a continuous value that is within the user-defined error bound to zero. This characteristic can cause non-negligible side-effects when implementing our proposed \techname. \textcolor{black}{Thus,} we must propose a modified version of SZ lossy compression to overcome this issue for our use case.
Second, since we plan to use an error-bounded lossy compressor, a strictly controlled compression error would be introduced to the activation data.
In order to maintain the training accuracy curve with a minimum impact to the performance and final model accuracy, \emph{we must understand how the introduced error would propagate through the whole training process.} In other words, we must theoretically and/or experimentally analyze the error propagation, which is challenging. To the best of our knowledge, there is no prior investigation on this.
Third, once we understand the connection between the controlled error and training accuracy, how to balance the compression ratio and accuracy degradation in a fine granularity is also challenging. 
In other words, a more aggressive compression can provide a higher compression ratio but also introduces more errors to activation data, which may significantly degrade the final model accuracy or training performance (cannot converge). 
Thus, we must find a balance to offer as high a compression ratio as possible to different layers across different iterations while maintaining minimal impact to the accuracy.
\section{Design Methodology}
\label{sec:design}

In this section, we describe the overall design of our proposed lossy compression supported CNN training framework \techname and analyze the performance overhead.

Our proposed memory-efficient framework \techname is shown in Figure~\ref{fig:fig-41-1}. We iteratively repeat the process shown in the figure for each convolutional layer in every iteration. \techname mainly includes four phases, shown in Figure~\ref{fig:fig-41-1} from left to right: (1) parameter collection of current training status for adaptive compression, (2) gradient assessment to determine the maximum acceptable gradient error, (3) estimation of compression configuration (e.g, absolute error bound), and (4) compression/decompression of activation data with our modified cuSZ. Note that the analysis of our error-bound control scheme for lossy compression of activation data that supports the \techname  design will be presented in Section~\ref{sec:analysis}. 

\subsection{Parameter Collection}
\label{sec:3.1}

First, we collect the parameters of the current training status for the following adjustment of lossy compression configurations. 
\techname mainly collects two types of parameters: (1) offline parameters in CNN architecture, and (2) semi-online parameters including activation data samples, gradient, and momentum.

First of all, we collect multiple static parameters including batch size, activation data size of each convolutional layer, and the size of its output layer. We need these parameters because they affect the number of elements considered into each value in the gradient and hence affect the standard deviation $\sigma$ in its normal error distribution, which will further impact the validation accuracy curve during training if introduced excessive error too.
It would also help the framework collects corresponding semi-online parameters.

For the semi-online parameters, we collect the sparsity of activation data and its average gradient of the loss in backpropagation to estimate how the compression error would propagate from the activation data to the gradient.
For the gradient, we compute the average value of its momentum. 
Note that in many DNN training frameworks such as Caffe \cite{jia2014caffe} and TensorFlow \cite{abadi2016tensorflow}, momentum is naturally supported and activated, so it can be easily accessed. The data collection phase is shown as the dashed thin arrows in Figure~\ref{fig:fig-41-1}.

Moreover, an active factor $W$ needs to be set at the beginning of training process to adjust the overall activeness in \techname.
$W$ is used to determine the activeness of our parameter extraction. 
We only extract semi-online parameters every $W$ iterations to reduce the computation overhead and improve the overall training performance. 
Based on our experiment, 
these parameters vary relatively slowly during training 
(i.e., the model would not change dramatically with reasonable learning rates in a short time). 
Thus, we only need to estimate the error impact in a fixed iteration interval in \techname. 
In this paper, we set $W$ to 1000 as default, which provides high accuracy and low overhead in our evaluation.
\textcolor{black}{
However, \techname would reduce $W$ by half if it determines that the maximum error-bound change of all layers between $W$ exceeds $2\times$, and reset $W$ to default when the error-bound settles to reduce the optimization time. We note that this technique is only effective for models that evolve rapidly during training.}
\textcolor{black}{Also, note that we still use decompressed data during training to collect these parameters, and the collected parameters from one period of iterations only affect during the following optimization computation.}

\subsection{Gradient Assessment}

Next, we estimate the limit of the gradient error that would result in little or no accuracy loss to the validation accuracy curve during training,
as shown in Figure~\ref{fig:fig-41-1}. 
Even with the help of the offset from momentum, we still want to keep the gradient of each iteration as close as possible to the original one. Based on our analysis (will be discussed in Section~\ref{sub:exp-analysis}), we need to determine the acceptable standard deviation $\sigma$ in the gradient error distribution that minimizes the impact on the overall validation accuracy curve during training.
We use 1\% as the acceptable error rate based on our empirical study (will be shown in Section~\ref{sec:evaluation:errorimpact}), i.e., the $\sigma$ in the momentum error model needs to be:
\begin{align}
    \sigma = 0.01 M_{Average},
    \label{equ-7} 
\end{align}
where $M_{Average}$ is the average value of the momentum. Note that here we use the average value instead of the modulus length of the momentum because we focus on each individual value of the gradient and the average value is more representative. 
The average value of the momentum can be considered as the average value of the gradient over a short time period based on the following equation:
\begin{align}
    M_t = \alpha G_{t-1}+\beta G_t,
    \label{equ-new-1} 
\end{align}
where $M_t$ is the momentum and $G_t$ is the gradient at iteration $t$.
We monitor the momentum by using the API provided by training framework (e.g., \texttt{MomentumOptimizer} in TensorFlow) and calculate its average value using simple matrix operations.
Similarly, based on our experiment, the average value of gradient does not tend to vary dramatically in a short time period during training.

\subsection{Activation Assessment}

After that, we dynamically configure the lossy compression for activation data based on the gradient assessment in the previous phase and the collected parameters as shown in Figure~\ref{fig:fig-41-1}. Based on our analysis (to be performed in Section \ref{sub:theory-analysis}), we need $\sigma$ (from gradient error model), $R$ (sparsity of activation data), $\bar{L}$ (average value of current loss), and $N$ (batch size) to determine the acceptable error bound for compressing the activation data at the current layer in order to satisfy the gradient error limit proposed in the previous phase. We simplify our estimator as below:
\begin{equation}
    eb = \frac{\sigma}{a \bar{L} \sqrt{N R}},
    \label{equ-8}
\end{equation}
where $eb$ is the absolute error bound for activation data with SZ lossy compressor, $\sigma$ describes the acceptable error distribution in the gradient, $a$ is the empirical coefficient, $\bar{L}$ is the average value of  the current layer's loss, $N$ is the batch size, and $R$ is the sparsity ratio of activation data.
Note that our technique can be applied to any non-momentum-based training, which only needs to monitor the “hidden” momentum which can be derived by gradient via simple matrix operations.

\subsection{Optimized Adaptive Compression}
\label{sub:comp}

In the last phase, we deploy the lossy compression with our optimized configuration to the corresponding convolutional layers. We also monitor the compression ratio for analysis.
\textcolor{black}{Note that we compress the activation data of each convolutional layer right after its forward pass.}
We \textcolor{black}{then} decompress the compressed activation data in the backpropagation when needed. 
\textcolor{black}{Also, we only perform the adaptive analysis in every $W$ iterations from Section~\ref{sec:3.1} to minimize the analysis overhead.}
\textcolor{black}{For Pooling, Normalization, and Activation layers, we use the recomputation technique to reduce their memory consumption since they take a large proportion of the memory consumption ($\sim60\%$ in the four tested models) with little recomputation overhead.
The other layers do not consume a noticeable amount of memory, so they are processed as is.}

\textcolor{black}{At the beginning of each training, the training dataset is unknown, and there is no collected semi-online parameter. \techname trains the model with original batch size to guarantee the memory constraint for the first $W$ iterations (negligible to the entire training process) to collect parameters.
After that, \techname starts to dynamically adjust the batch size based on the previous compression ratio and control the remaining memory space under the following optimizations: (1) choosing the batch size of $2^k$ to stabilize the performance, where $k>0$ is an integer; (2) defining a maximum batch size to avoid unnecessary scaling for smaller models; and (3) reserving 5\% of the total available GPU memory when calculating the capable batch size to avoid overflow caused by unexpectedly low compression ratio.
For the extremely low compression ratio case, the compressed data will be evacuated to CPU with a certain communication overhead. However, in our evaluation, this rarely happens with the 5\% reserved memory space 
and thus results in a negligible overhead to the system.
We also note that the optimized batch size is almost constant thanks to the batch size setting of $2^k$, and the compression ratio is relatively stable on our test models. We will try to provide a dynamic batch size 
in future work.}

\textcolor{black}{T}hrough analysis in Section~\ref{sub:theory-analysis} and evaluation in Section~\ref{sec:evaluation:errorimpact}, we identify that the current version of SZ algorithm cannot always reconstruct continuous zeros as \textcolor{black}{an} exact zero but introduce a small shift (within the error bound) to those continuous zeros. which can eventually cause gradient explosion and an untrainable model (showed in Figure \ref{fig:fig-5-sz}).
Thus, we need to force zeros in activation data to remain unchanged in the compression algorithm to maximize the performance of \techname, 
as discussed in Section~\ref{subsec:challenge}. 
To solve this issue, we propose an improved version of cuSZ algorithm~\cite{tian2020cusz} to handle the case of compressing continuous zeros.
Specifically, we add a filter to the decompression process to re-zero those values within the error bound. 
Every reconstructed value that has the distance to zero within the error bound would be decompressed as zero. 
Another solution is to leverage those values into zeros during the first quantization step of cuSZ's dual-quant mechanism when compressing to grantee the reconstructed values remain zeros.
Compared to the second solution, re-zero those values during decompression means we can still use these non-zero reconstructed values for their following points' prediction instead of using zero during compression, which can ensure our compression ratio would not be affected. 
\textcolor{black}{
By doing so, we will inevitably flush some small values into zeros, but they only take a small proportion and contribute little when calculating the gradient. Thus, it is not worth adding extra overhead to distinguish these decompressed zeros from real zeros.}
\section{Compression Error Impact Analysis}
\label{sec:analysis}

In this section, we present the analytical support of our proposed training framework---analyzing compression error propagation (1) from activation data to gradient and (2) from gradient to validation accuracy curve during training for convolutional layers. 

\subsection{Modeling Compression Error}
\label{subsec:model}

cuSZ \cite{tian2020cusz} is a prediction-based error-bounded lossy compressor for floating-point data on GPUs.
It first uses a dual-quantization technique to quantize the floating-point input data based on user-set error bound. Then, it applies Lorenzo-based predictor \cite{ibarria2003out} to efficiently predict the value of each data point based on its neighboring points.  After that, a quantization code (integer) is generated for each value. Finally, a customized Huffman coding is applied to all the quantization codes. Similar to the original SZ, the error introduced to the input data after decompression usually forms a uniform distribution. This is mainly because of the linear-scaling quantization technique adopted. We refer readers to \cite{lindstrom2017error} for more details about the error distribution of SZ from a statistical perspective. 

\begin{figure}[]
    \centering
    \includegraphics[width=0.8\linewidth]{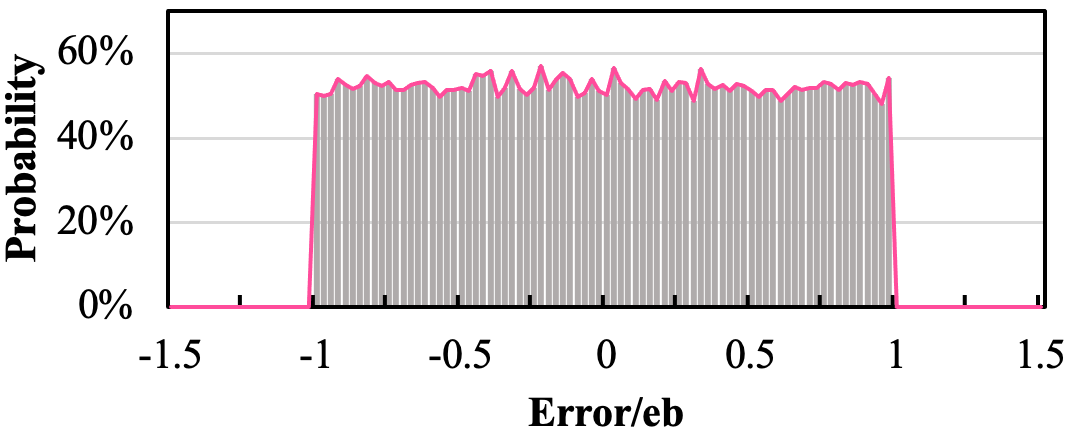}
    \vspace{-3mm}
    \caption{An example of compression error distribution of activation data compressed by cuSZ error-bounded lossy compression with absolute error bound $eb = 10^{-4}$.}
    \label{fig:fig-31-1}
\end{figure}

Figure~\ref{fig:fig-31-1} illustrates an example of compression error distribution when compressing/decompressing the activation data (generated by the 5th convolutional layer of AlexNet \cite{krizhevsky2012imagenet} with the ImageNet dataset) by cuSZ. 
Note that we plot the error distribution every 50 iterations, and observe that all the error distributions are quite similar and follow uniform distribution, which is consistent with the conclusion drawn in the prior work \cite{lindstrom2017error}. 
\textcolor{black}{In fact, based on the evaluation in Section~\ref{sec:evaluation}, \techname compresses the activation data with a compression ratio less than $20\times$, where the error distribution is uniform in theory with SZ compression~\cite{jin2021novel}.}
Thus, we propose to use the uniformly distributed error model to perform analysis and an error-injection-based approach to demonstrate the effectiveness of our theoretical derivation in this section: 
\begin{align}
    e \sim U[-eb, +eb] = \begin{cases}
    \frac{1}{2eb}, &-eb \leq x \leq eb, \\
    0, &\text{otherwise,}
    \end{cases}
\end{align}
where $eb$ is the user-defined absolute error bound for SZ lossy compressor.
Note that for the purpose of our preformed analysis, we inject the error, rather than actually compressing activation data, to demonstrate how uniformly distributed error propagates from the activation data to the gradient and then to the whole training process. We will use actual compression/decompression in our following evaluation.

\subsection{Modeling Error Impact on Gradient}
\label{sub:theory-analysis}

Next, we theoretically derive how error propagates from activation data to gradient and provide experimental proof based on statistical analysis using error injection.

\begin{figure}[]
    \centering
    \includegraphics[width=0.9\linewidth]{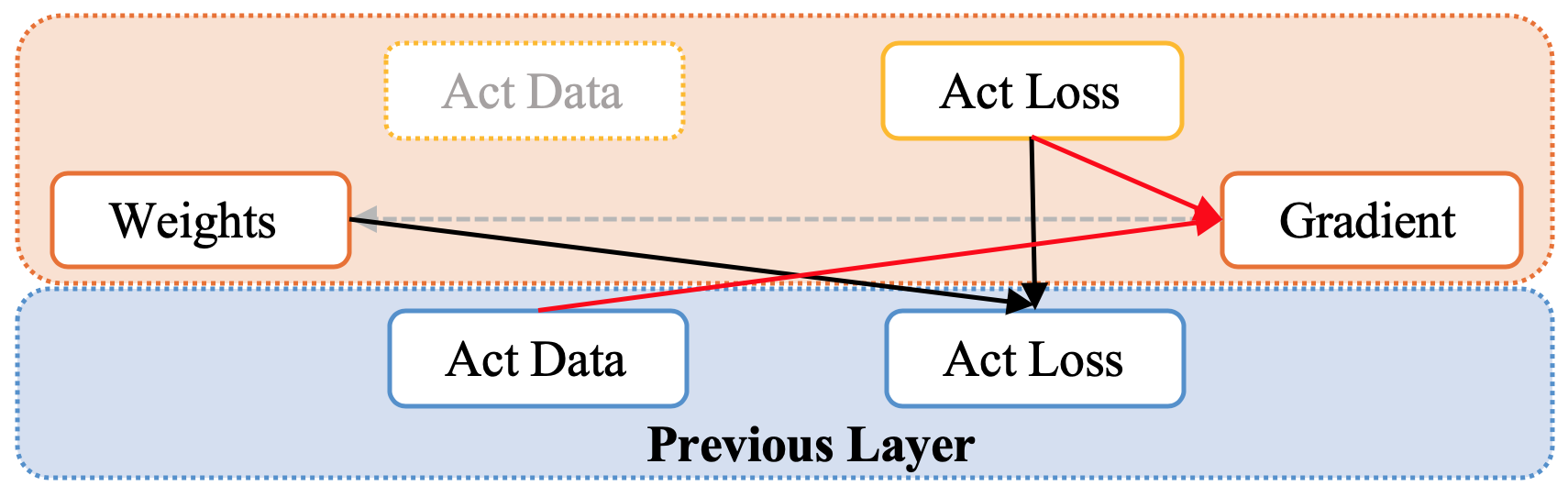}
    \vspace{-2mm}
    \caption{Data dependencies in one convolutional layer during backpropagation.}
    \label{fig:fig-32-3}
    \vspace{-2mm}
\end{figure}

As aforementioned (shown in Figure~\ref{fig:fig-1}), the compressed activation data needs to be decompressed when the backpropagation reaches the corresponding layer. During the backpropagation, each layer computes the gradient to update the weights and the gradient of the loss to be propagated to the previous layer (backpropagation). 
As shown in Figure~\ref{fig:fig-32-3}, on the one hand, the gradient of the loss of activation data for the previous layer only depends on the current layer's gradient of the loss and weight. 
On the other hand, the gradient depends not only on the gradient of the loss of the current layer but also on the activation data.
We note that errors introduced in activation data do not pass across layers along with loss function.
In conclusion, in order to understand the impact of compressing the activation data, we must first understand how compression error introduced to the activation data would propagate to the gradient.

\begin{figure}[]
	\begin{subfigure}{\linewidth}\centering
	    \includegraphics[width=0.8\linewidth]{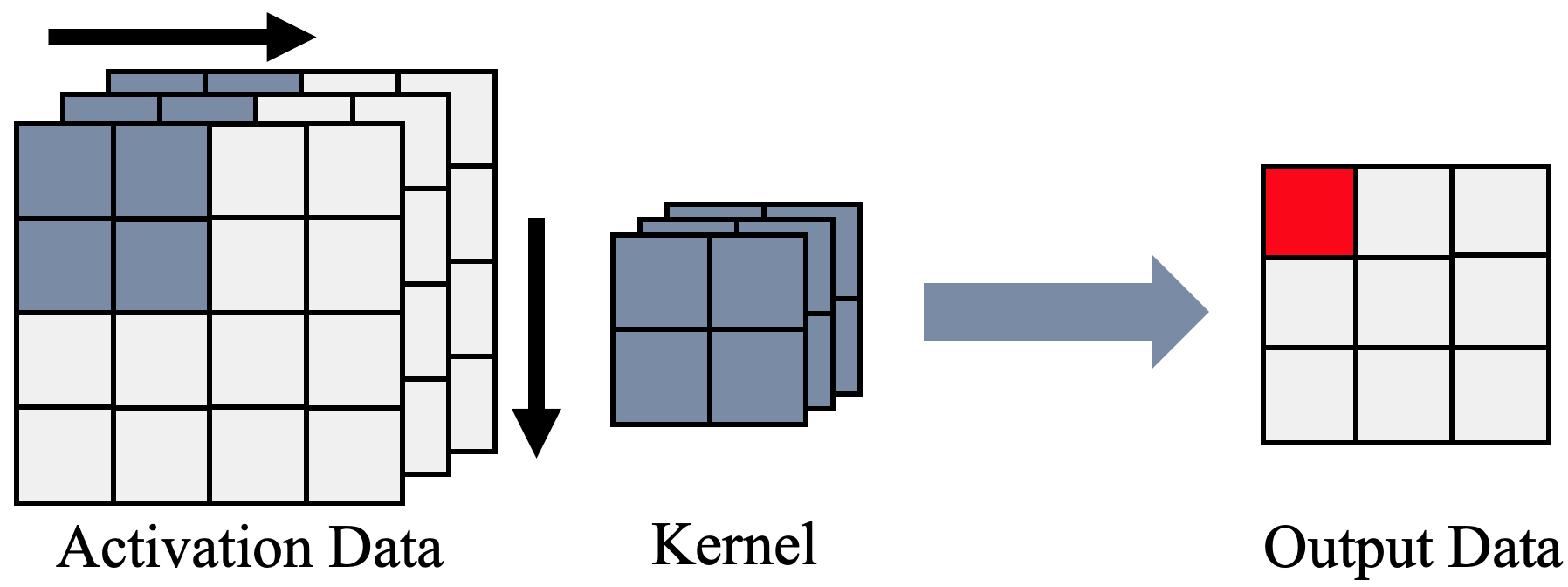}
	    \caption{\footnotesize Forward pass}\label{fig:subfigure_dependency_forward}
	\end{subfigure}
	\begin{subfigure}{\linewidth}\centering
	    \includegraphics[width=0.8\linewidth]{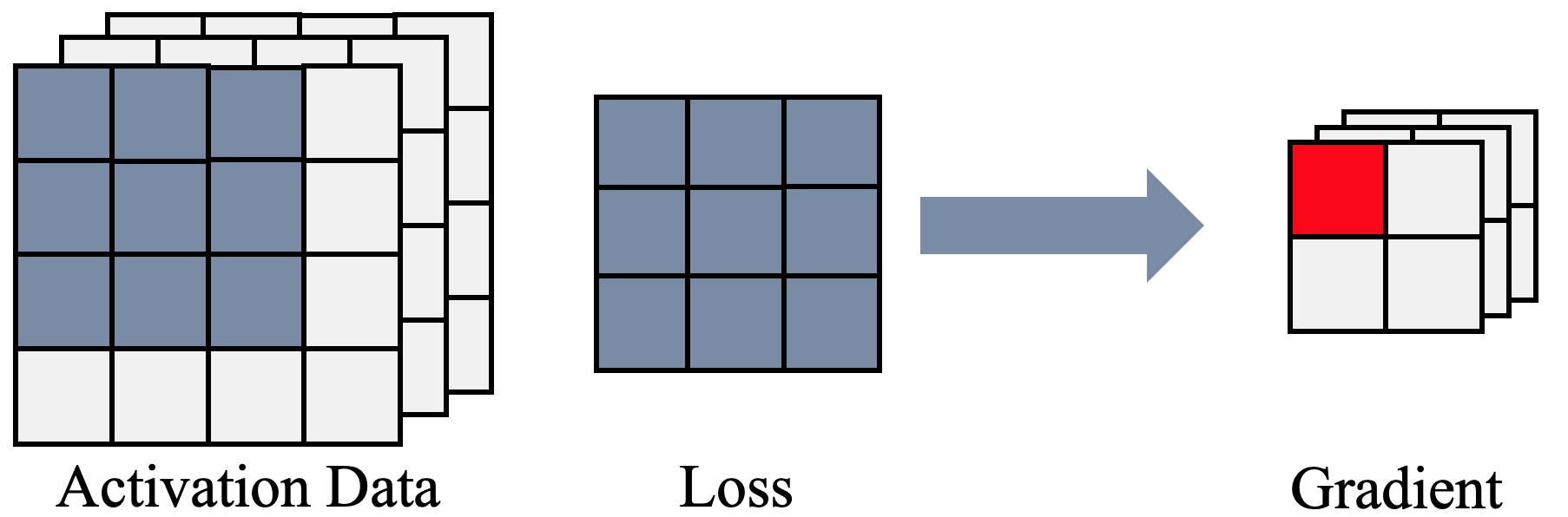}
	    \caption{\footnotesize Backward pass}\label{fig:subfigure_dependency_backward}
	\end{subfigure}
\vspace{-4mm}
\caption{Forward and backward computation in a sample convolutional layer. Kernel size is 2$\times$2, stride is 1, input channel count is 3, and output channel count is 1. Batch size is on another dimension and shown as 1.}
\label{fig:fig-32-4}
\end{figure}

In the forward pass, multiple kernels perform convolutions on the input activation data. 
As shown in Figure~\ref{fig:subfigure_dependency_forward}, the kernel is performed on the activation data (marked in brown) and generates the output value as shown in red. 
Similar to the forward pass, the backward pass reverses the computation, where the parameter's gradient is computed based on the gradient of the loss (with the same dimension of the output data in the forward pass) and the original data in the kernel, as shown in Figure~\ref{fig:subfigure_dependency_backward}.
Similarly, the activation data and the gradient of the loss calculate the gradient value (as shown in red). More specifically, this value is computed as
\begin{align}
\textstyle
    G_{k,i} = \sum_{i=0}^{n} A_{k,i^{'}} {\times} L_{i},
    \label{equ-1}
\end{align}
where $G$ is the gradient, $A$ is the activation data, $L$ is the gradient of the loss, $k$ is the current channel, $i$ is the value index of the channel, $n$ is the number of values in the gradient of the loss matrix, $i^{'}$ is the corresponding index of activation data to the gradient of the loss matrix. Note that for simplicity, we ignore all $i$ on the right side of Equation (\ref{equ-1}). 
Note that here, if the number of output channels is greater than 1, which is true for most convolutional layers, the same process will be used for multiple kernels, as shown in Figure~\ref{fig:subfigure_dependency_backward}, and the above formula still holds in this case.

Based on our analysis in Section \ref{subsec:model}, the error introduced to the activation data is uniformly distributed.
Thus, in the backpropagation, we can have
\begin{align}
    G^{'}_{k,i} &= \textstyle\sum_{i=0}^{n} A^{'}_{k,i^{'}} {\times} L_{i}, \label{equ-2} \\
    A^{'}_{k,j} &= A_{k,j} + e, e \sim U[-eb, +eb], \notag
\end{align}
where $A^{'}$ is the decompressed activation data,
$G^{'}$ is the gradient altered by the compression error, $e$ is the error, and $eb$ is the user-set absolute error bound. After a simple \textcolor{black}{transform}, we can have
\begin{align}
    G^{'}_{k,i} &= \textstyle\sum_{i=0}^{n}  A_{k,i^{'}} {\times} L_{i} + \sum_{i=0}^{n} e_i {\times} L_{i} = G_{k,i} + E, \label{equ-3} \\
    E &= \textstyle\sum_{i=0}^{n} e_i {\times} L_{i},\notag
\end{align}
where $E$ is the gradient error. 

Although it is not possible to calculate or predict the exact value of every element $E$, we can predict its distribution based on our previous assumption. 
We also note that the batch size of a typical neural network is usually relatively large during the training process, such as 256, since a larger batch size results in higher training performance in general.
As a result, the final gradient for updating weights can be computed as follows. 
\begin{align}
    G^{'}_{final} &= Average(G^{'}_0, G^{'}_1, ... , G^{'}_N), \notag\\
    E_{final} &= Average(E_0, E_1, ... , E_N) \label{equ-4} \\
    &= \textstyle\sum_{i=0}^{n} \sum_{j=0}^{N} e_{j,i} {\times} L_{j,i}, \notag \\
    e &\sim U[-eb, +eb], \notag
\end{align}
where $N$ is the batch size. 
Note that all $e$s are independently and uniformly distributed as discussed in Section \ref{subsec:model}. Although $L$ can be related to each other in the same batch, they are still independently distributed across different batches. 
According to Central Limit Theorem~\cite{wiki}, the sum of a series of independent random variables with the same distribution follows a normal distribution,
which means the error distribution of the gradient can be expected to be normally distributed. 
We identify that the distribution of the gradient of the loss $L$ for one input (i.e., one image) is highly concentrated in zero, where the highest value in the gradient of the loss is usually much larger than the average of $L$. Thus, we can simplify Equation \ref{equ-4} to 
\begin{align}
    E_{final} & \approx \textstyle\sum_{i=0}^{N} e_{i} {\times} L_{max, i}, \label{equ-8} \\
    e &\sim U[-eb, +eb], \notag
\end{align}
where $L_{max}$ is the maximum value in the gradient of the loss for each input alone. To reduce the complexity, we can greatly improve the performance with fewer parameters that need to be collected.

\begin{figure}[]
    \centering
    \begin{subfigure}{\linewidth}
    \centering
    \includegraphics[width=0.95\linewidth]{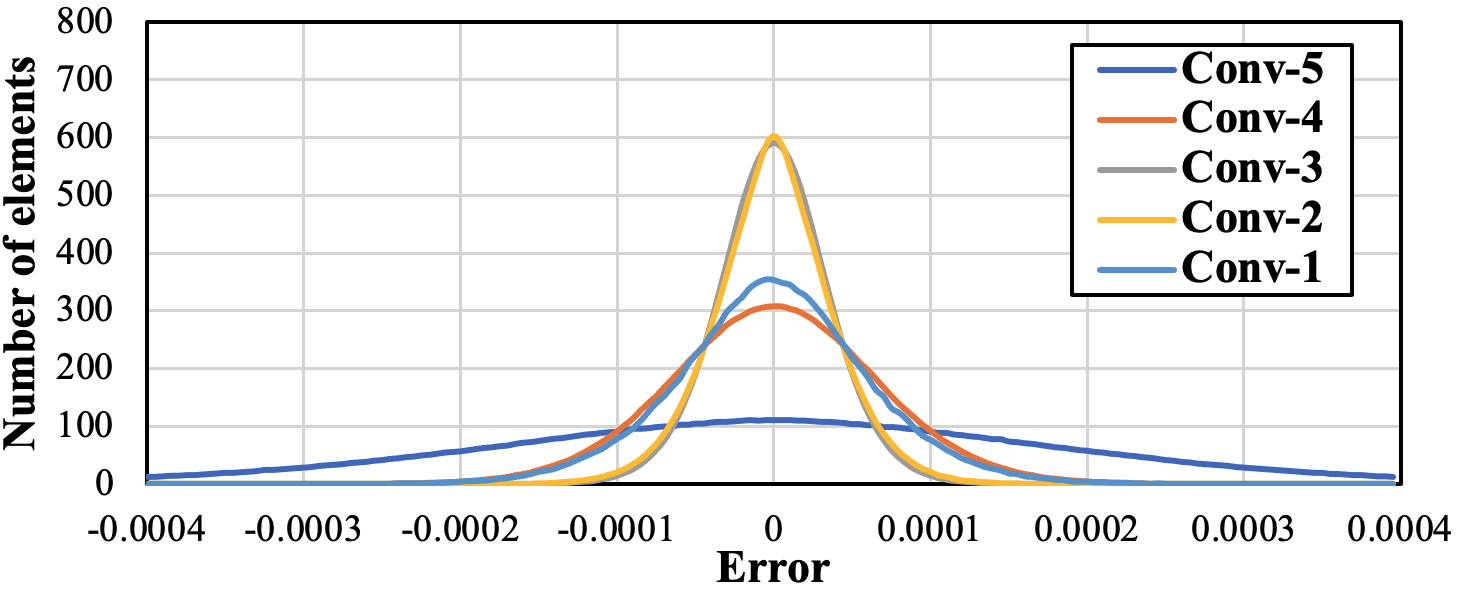}
    \caption{\footnotesize Original zeros have been compressed}
    \label{fig:fig-32-1}
    \end{subfigure}

    \begin{subfigure}{\linewidth}
    \centering
    \includegraphics[width=0.95\linewidth]{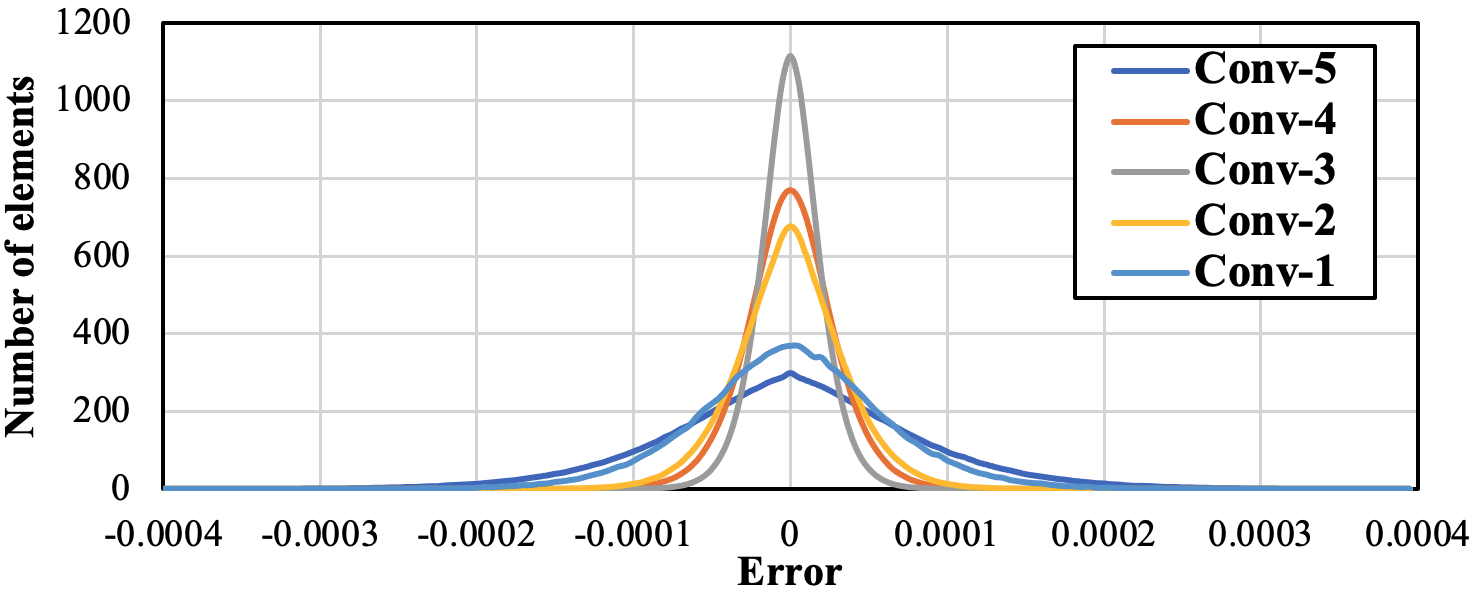}
    \caption{\footnotesize Original zeros remain zero}
    \label{fig:fig-32-2}
    \end{subfigure}
    \vspace{-2mm}
    \caption{Distributions of gradient error when injecting modeled compression error to activation data. Note that the number of elements is normalized to ensure different layers under the same scale. Data is collected every 100 iterations.}
    \label{fig:fig-32}
\end{figure}

Next, we inject error (to simulate the compression error) to the activation data in convolutional layers based on our error model discussed in Section \ref{subsec:model}. Then, we collect the error of the gradient in the backpropagation. 

Figure~\ref{fig:fig-32-1} illustrates the normalized error distribution of gradients collected from different layers, all of which follow the normal distribution as expected. 
In fact, by calculating the percentage of the area within $\pm\sigma$ of each curve, we can get a value close to 68.2\%, which confirms our theoretical derivation. 
Then, we need to figure out how to predict $\sigma$ before compression in order to calculate the desired and acceptable error bound for each layer's activation data.

First, we note that $\sigma$ is highly related to the number of elements that are combined together. In general, more elements result in larger $\sigma$, \textcolor{black}{and} vice versa.
A $2\times$ increase of elements results in $\sqrt{2}\times$ increase of $\sigma$, which means that more uncertainties have been added to the system. 
Second, $\sigma$ is also related to the value scale, in this case, the average of the gradient of the loss $L$ at the current layer. 
Note that it is not necessary to compute the average value of the gradient of the loss in every iteration. Instead, we can compute it every $W$ iterations to reduce the computation overhead, since this value is relatively stable in a fixed period of time. 
Based on these two parameters, we can estimate $\sigma$ by the following equation:
\begin{align}
    \sigma \approx a \bar L \sqrt{N} eb,
    \label{equ-5}
\end{align}
where $L$ is the gradient of the loss matrix, $N$ is the batch size, and $a$ is an empirical coefficient. Note that this coefficient $a$ is unchanged for different neural networks because it is essentially a simplified value of the previous equation.

Finally, we note that there is a notable fraction of activation data to be zeros, however, our above analysis so far does not cover it. 
When compressing a series of continuous zeros, the original cuSZ may change them into a continuous small value within the user-defined error bound instead of exactly zero. 
Because the reconstructed continuous small values are the same value due to the Lorenzo-based predictor in SZ, this can cause the gradient computation to have nonnegligible offset that eventually can cause gradient exploding problems.
To solve this issue, we propose an improved version of cuSZ to enhance the compression on continuous zeros as discussed in Section~\ref{sec:design}.
This means no error would be introduced when facing continuous zeros. 
Moreover, in some cases, the activation data may contain many zeros due to the activation function layer (such as the ReLU layer) before the current convolutional layer. If this happens, the activation data of convolutional layer can be quickly recomputed through the activation function instead of being saved, which will essentially erase the negative values to zeros. 
Since lossy compression such as cuSZ is unlikely to change the sign of the activation data value, these data will remain at zero.

Figure~\ref{fig:fig-32-2} shows the distribution of the gradient error after we inject the error into the activation data (maintaining zeros unchanged). Compared with Figure~\ref{fig:fig-32-1}, we can observe a decrease of $\sigma$, but it still holds a normal distribution. This decrease is partly because of the reduction of the number of elements in Equation~\ref{equ-4}, since those zeros do not have any error.
In these cases, we can revise the prediction of $\sigma$ accordingly by the following equation:
\begin{align}
    \sigma^{'} = \sigma \sqrt{R},
    \label{equ-9}
\end{align}
where $R$ is the ratio of non-zero elements percentage in the activation data. Again, in practice, we do not need to compute this ratio every iteration but every $W$ iterations, since this ratio is relatively stable in a fixed period of time.

\subsection{Error Impact Analysis}

\label{sub:exp-analysis}

Finally, we discuss the error propagated from gradient to overall validation accuracy curve during training using an experimental analysis. 
Our goal is to identify the maximum acceptable gradient error that would cause little or no validation accuracy loss.
According to our performed analysis in Section \ref{sub:theory-analysis}, the gradient error can be modeled as a normally distributed error. 

In this subsection, we follow the same strategy used in the last subsection to inject error to the gradient that follows our error model and perform the analysis and evaluation. 
It is worth noting that similar to many existing studies (e.g., CNN model pruning \cite{han2015deep}, compression \cite{jin2019deepsz}, mixed-precision training~\cite{micikevicius2017mixed}), 
our hypothesis is that \textit{the accuracy loss caused by the errors added to a given convolutional layer is not noticeably amplified by its following layers}.
Other existing study~\cite{ying2019overview} also points out that adding noise to the training data can even provide a regularization effect that can help improve the training performance from overfitting. 
\textcolor{black}{Our introduced error is slightly different from purely adding noise to training data, but rather it can be considered as noise with a uniform distribution on the nonzero activation data, which only affects the gradient computation during the backpropagation phase and precludes any error propagation across layers. This might degrade the training performance due to the alternated gradient update.}
\textcolor{black}{However, as pointed out by Lin \textit{et al.}~\cite{lin2017deep}, gradient can be considerably approximated 
until the training performance is affected. In addition, such noise can potentially help training get rid of local minima, especially for models with rough loss surfaces, while providing similar convergence speeds for models with smooth loss surfaces~\cite{li2017visualizing}.}

Momentum has been widely adopted in most neural network training \cite{dong2018boosting}, which can be used for alleviating the impact of the gradient error~\cite{ruder2016overview}.
Actually, in order to update the weights, it is based not only on the gradient computed from the current iteration but also on the momentum.
In other words, both the gradient and the momentum (with the same dimension as the weights and gradient) take up a portion of the updated data for weights. 
Thus, it is critical to maintaining an accurate momentum vector (similar to the error-free one) to guide the weight update. 
While thanks to the normally distributed gradient error, which is centralized and symmetric in the original direction, the momentum error is relatively low compared to the gradient error. 
Therefore, this does help the training towards the correct direction---even a few iterations may generate the undesired gradient, they can be offset quickly through the momentum based on the estimated gradient error.
Similarly, other optimization algorithms such as Adagrad and Adam~\cite{ruder2016overview} also follow a similar principle and can be benefited from \techname.

\section{Experimental Evaluation}
\label{sec:evaluation}

In this section, we evaluate \techname from four aspects, including (1) evaluation of compression error impact on gradient, (2) evaluation of error propagation from gradient to training curve, (3) comparison between \techname and the state-of-the-art method, and (4) performance evaluation with multiple state-of-the-art GPUs.
\textcolor{black}{We manually set the batch size of \techname for observation and variable control purposes unless specified.}

\subsection{Experimental Setup}

Our evaluation is conducted with Caffe \cite{jia2014caffe} and Tensorflow 1.15~\cite{abadi2016tensorflow}. We choose Caffe for a single-node experiment due to its easy-to-modify architecture and choose Tensorflow for multi-node evaluation due to its being widely used in the research. 
Our experiment platform is the Longhorn system \cite{longhorn} at TACC and the Bridge-2 system \cite{bridge2} at PSC, of which each GPU node is equipped with 4/8 NVIDIA Tesla V100 GPUs \cite{nv100} per node. Our evaluation dataset is the ImageNet-2012 \cite{krizhevsky2012imagenet} and Stanford Dogs Dataset \cite{KhoslaYaoJayadevaprakashFeiFei_FGVC2011}. 
The CNN models used for image classification include AlexNet \cite{krizhevsky2012imagenet}, VGG-16 \cite{simonyan2014very}, ResNet-18, ResNet-50 \cite{he2016deep}, and EfficientNet~\cite{tan2019efficientnet}.

\subsection{Error Impact Evaluation}
\label{sec:evaluation:errorimpact}

\begin{figure}[]
    \centering
    \includegraphics[width=0.55\linewidth]{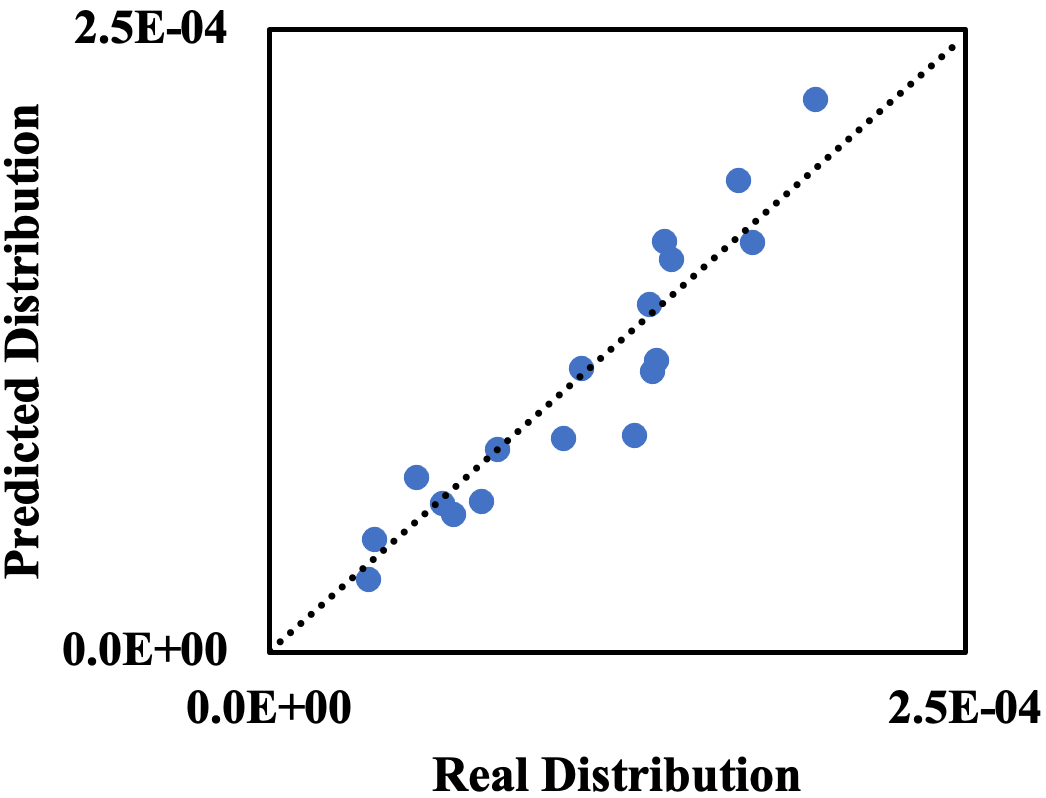}
    \vspace{-3mm}
    \caption{Comparison of standard deviation $\sigma$ of gradient error (caused by compression error introduced to activation data) from measured distribution and from predicted distribution (based on our theoretical analysis).}
    \vspace{-2mm}
    \label{fig:fig-5-dis}
\end{figure}

\begin{figure}[]
    \includegraphics[width=0.9\linewidth]{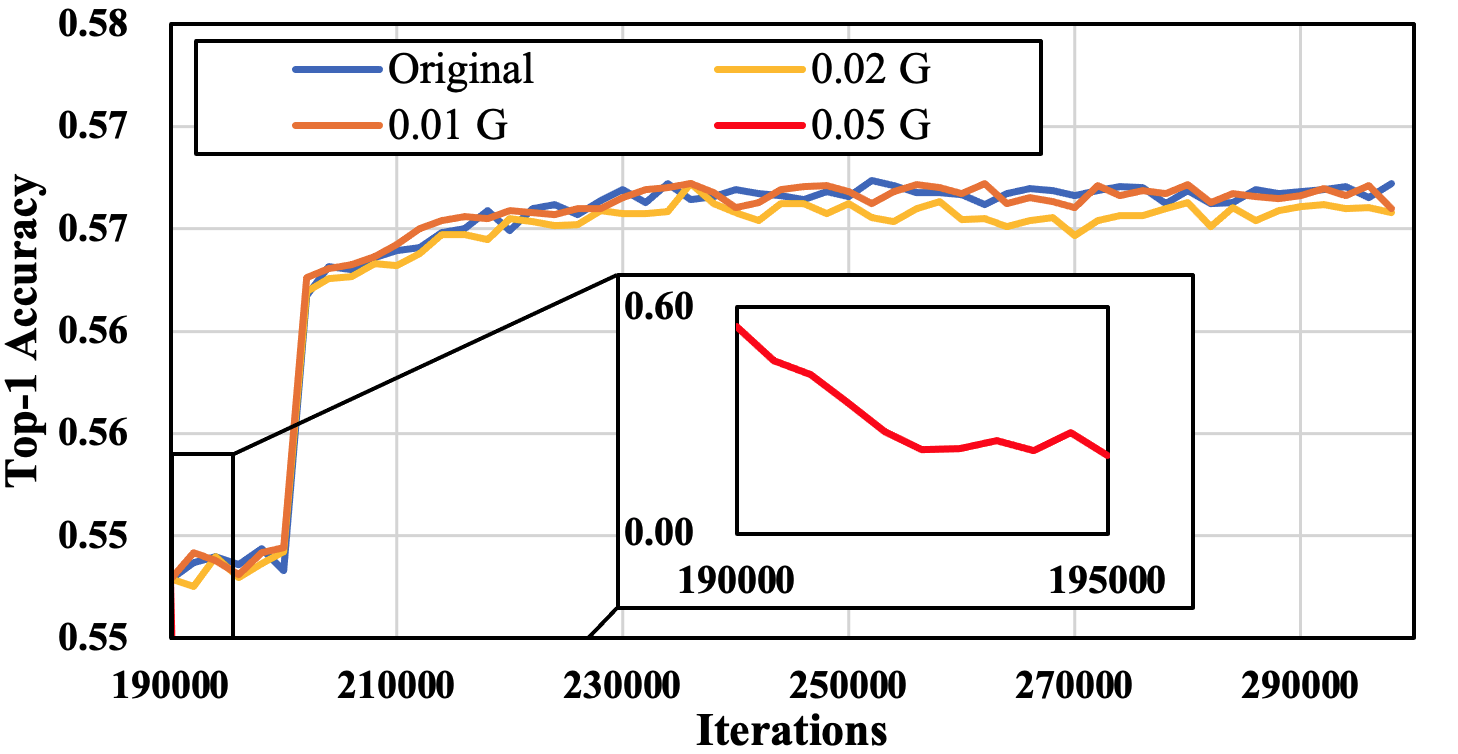}
    \vspace{-3mm}
    \caption{\textcolor{black}{Validation accuracy training curve started from itration 190,000. Learning rate updated at iteration 200,000} Evaluation on different fractions of error introduced to the gradient. $G$ represents for average value of gradient with coefficient value used in \techname.}
    \label{fig:fig-5-sigma} 
\end{figure} 

First, we evaluate our proposed theoretical analysis in Section \ref{sub:theory-analysis}.
Based on Equations~\ref{equ-5} and \ref{equ-9}, we can estimate $\sigma$ which stands for how an error is distributed in the gradient. 
After implementing our estimation, we identify that coefficient $a$ in Equation~\ref{equ-5} is $0.32$ based on our experiment. 
This is reasonable because if we consider the extreme condition that the batch size $N = 1$, the error distribution in the gradient will be the same as the SZ lossy compression to uniformly distribute and result in $a=1/3$. 

We also evaluate our estimation on different layers of AlexNet and VGG-16 using the batch size of 256, as shown in Figure~\ref{fig:fig-5-dis}. 
We can clearly observe that the coefficient and how our estimated value aligns with the actual error distribution. 
This means that we can not only estimate the error propagation but also determine the error bound based on a given acceptable $\sigma$ error distribution.

Next, we evaluate the error impact from gradient to the overall training process in terms of validation accuracy, as discussed in Section~\ref{sub:exp-analysis}.
Since we target to cause little or no accuracy loss, we focus on the iterations close to the end of training in this evaluation, since the accuracy is harder to be increased when the training is close to the end. 
To reduce the training time and find an empirical solution for this specific analysis, we pre-train the model without \techname first and save the snapshot every epoch. Then, we perform our evaluation of error impact analysis using those snapshots from different iterations to demonstrate the effectiveness of \techname.

Figure~\ref{fig:fig-5-sigma} shows our experiment with AlexNet starting from the iteration of 190,000 with a batch size of 256.
Here smaller coefficient value means less error introduced to the gradient but potentially lower compression ratio.
On the one hand, we can observe from the zoomed-in subfigure that $\sigma = 0.05$ would result in an unacceptable error loss that cannot be eventually recovered. 
On the other hand, $\sigma = 0.02$ can provide better accuracy and a higher compression ratio, but it does affect the accuracy a bit in some cases. 
Thus, considering that our goal is a general solution for convolutional layers, we eventually choose $\sigma = 0.01$ as default in \techname; in other words, the target $\sigma$ is 1\% of the average of gradient.
In fact, we evaluate 0.1\%$\sim$5\% for $\sigma$ and choose the best one (i.e., 1\%) to minimize the accuracy impact.

\begin{figure}[]
    \includegraphics[width=0.87\linewidth]{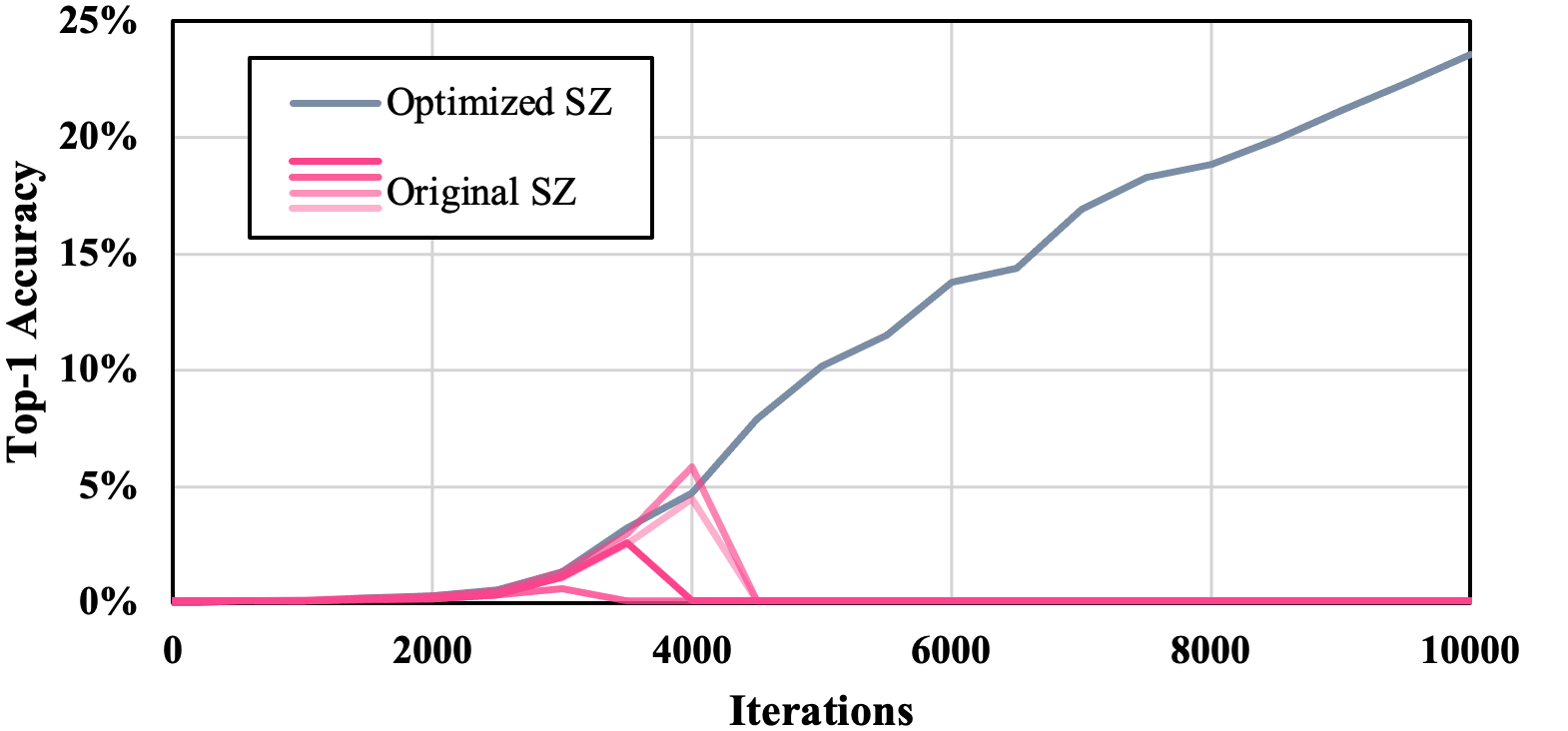}
    \vspace{-3mm}
    \caption{\textcolor{black}{Validation accuracy training curve at the start of training.} Comparison of validation accuracy (during training) using original SZ and our optimized SZ in \techname on AlexNet (batch size = 256). The evaluation of the original SZ is conducted four times (with four curves).}
    \label{fig:fig-5-sz} 
\end{figure}

Last but not the least, we also evaluate the effectiveness of our proposed optimized SZ, by adding a filter to the decompression process to re-zero values that are closer to zero than the defined error bound in order to improve the performance for the reconstruction of contiguous zeros. As shown in Figure~\ref{fig:fig-5-sz}, we can observe that without our modification to the current version of SZ, the training process cannot even \textcolor{black}{sustains} through 5,000 iterations. This is because the original SZ would usually reconstruct continuous zeros to a continuous small value by design of the Lorenzo predictor (i.e., only ensuring the reconstructed values to be within the pre-defined error bound to the original values) and can cause gradient update to shift from original that eventually leads to gradient explosion.
Note that we perform the experiment with the original SZ four times, leading to crashes in four different iterations due to their different training initial states and the uncertainty of gradient explosion.
Figure~\ref{fig:fig-5-sz} illustrates that our optimized SZ can solve this issue and provide stable performance during the training process. 

\subsection{Memory Reduction Evaluation} 

\begin{figure}
    \centering
    \includegraphics[width=0.95\linewidth]{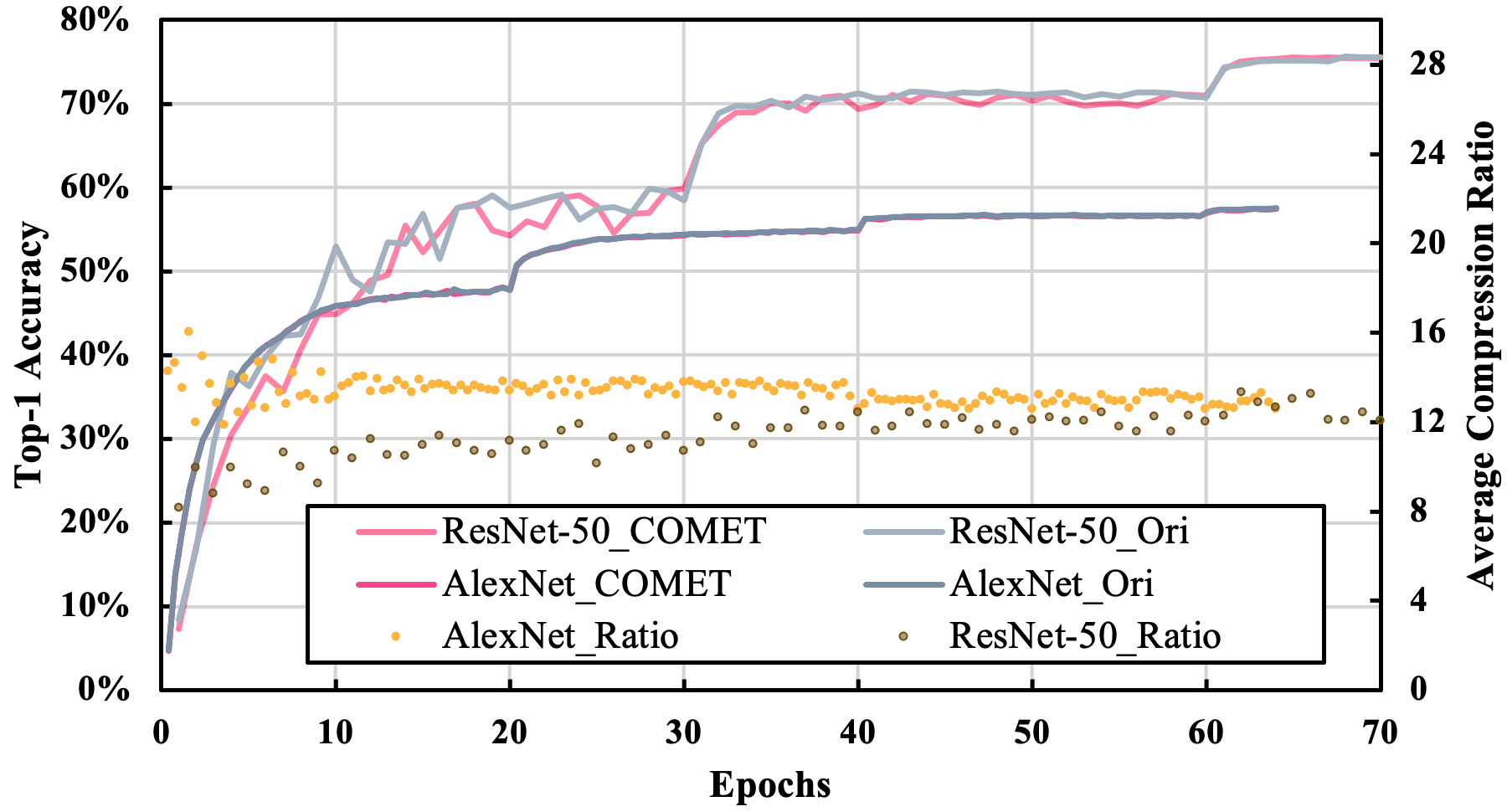}
    \vspace{-3mm}
    \caption{Comparison of validation accuracy between baseline training and \techname (batch size = 256). Lines represent the validation accuracy during training; dots represent the compression ratio of \techname on secondary y-axis.}
    \label{fig:fig-5-alexnet}
\end{figure}

We test \techname on various popular CNNs and evaluate its memory reduction capability. 
We use the original training approach of each model without memory reduction techniques (i.e., recomputation, migration, and compression) as the baseline.
Figure~\ref{fig:fig-5-alexnet} illustrates the result with AlexNet and ResNet-50. The black and red lines are the validation accuracy curves of the baseline training and \techname. 
We can observe that these two curves are very close to each other, meaning \techname does not \textcolor{black}{noticeably} affect the validation accuracy. We also illustrate the change of compression ratio to iteration in yellow dots. In the early stage of training, the compression ratio can be slightly unstable because of the relatively large change to the model itself. Note that the compression ratio will change slightly when the learning rate changes because the learning rate only matters when updating the gradient to the weights. 
Other than that, for some layers, although the average maximum loss of each input should be decreased and result in a higher error bound, the corresponding activation data value is actually increased. Thus, the compression ratio would not increase even with a higher error bound.
Moreover, we evaluate a static strategy for comparison: we estimate the error bound only once at the beginning of training (i.e., at iteration 1, 100, and 200) and keep using this error bound through the whole training process. This static strategy leads to a significant accuracy drop (i.e., top-1 accuracy of only 30.6\%, 36.7\%, and 31.0\% on AlexNet at 10 epochs, respectively) compared to the baseline training (i.e., 45.9\% at 10 epochs), which proves the necessity of adaptively configuring error bound. 

Table~\ref{tab:compress} shows the compression ratio of convolutional layers \textcolor{black}{and the overall peak memory usage} that \techname can provide \textcolor{black}{at the batch size of 128. ``Max batch'' means the maximum batch size that the baseline and \techname can run with on a single GPU with 16 GB memory.}
There is almost no accuracy loss or only little, with up to $0.31$\%.
\textcolor{black}{T}hanks to our careful control of compression error and thorough analysis and modeling of error impact. 
\techname can deliver a promising compression ratio without heavy efforts of fine-tuning any parameter for different models.
Overall, \techname can provide up to 13.5$\times$ compression ratio with little or no validation accuracy loss. 

\begin{table}[]
    \caption{Comparison of validation accuracy between baseline training and \techname; comparison of compression ratio between JPEG-ACT and \techname .}
    \vspace{-4mm}
    \label{tab:compress}
    \footnotesize{
\resizebox{0.99\columnwidth}{!}{%
\begin{tabular}{@{}rrrrrr|r@{}}
\toprule
\textbf{Neural Nets} & 
    \begin{tabular}{@{}r@{}} \bfseries Top-1 \\\bfseries Accuracy \end{tabular} & 
    \begin{tabular}{@{}r@{}} \bfseries \textcolor{black}{Peak} \\ \textcolor{black}{\bfseries Mem.} \end{tabular} & 
    \begin{tabular}{@{}r@{}} \bfseries \textcolor{black}{Max} \\ \textcolor{black}{\bfseries Batch} \end{tabular} & 
    \begin{tabular}{@{}r@{}} \bfseries Conv. \\ \bfseries Act. Size \end{tabular} & 
\textbf{COMET} &
    \begin{tabular}{@{}r@{}} \bfseries JPEG- \\\bfseries ACT \end{tabular} \\
\midrule
b. & 57.41\% & \textcolor{black}{2.17 GB} & \textcolor{black}{512} & 407 \makebox[1.5em][l]{MB} & \\
\textbf{AlexNet} c. & 57.42\% & \textcolor{black}{0.85 GB} & \textcolor{black}{2048} & \bfseries 30 \makebox[1.5em][l]{MB} & 13.5$\times$ & - \\
\cmidrule{2-5}
b. & 
	68.05\% & \textcolor{black}{17.29 GB} & \textcolor{black}{64} &
	6.91 \makebox[1.5em][l]{GB} & \\
\textbf{VGG-16} c. & 
	68.02\% & \textcolor{black}{5.04 GB} & \textcolor{black}{256} &
	\bfseries 0.62 \makebox[1.5em][l]{GB} & 
	11.1 $\times$ & - \\
\cmidrule{2-5}
b. & 
	67.57\% & \textcolor{black}{5.16 GB} & \textcolor{black}{256} &
	1.71 \makebox[1.5em][l]{GB} & 
	\\
\textbf{ResNet-18} c. & 
	67.43\% & \textcolor{black}{1.37 GB} & \textcolor{black}{1024} &
	\bfseries 0.16 \makebox[1.5em][l]{GB} & 
	10.7 $\times$ & 7.3 $\times$ \\
\cmidrule{2-5}
b. & 
	75.55\% & \textcolor{black}{15.57 GB} & \textcolor{black}{128} &
	5.14 \makebox[1.5em][l]{GB} & 
	\\
\textbf{ResNet-50} c. & 
	75.51\% & \textcolor{black}{4.40 GB} & \textcolor{black}{512} &
	\bfseries 0.46 \makebox[1.5em][l]{GB} & 
	11.0 $\times$ & 6.0 $\times$ \\
\bottomrule
\end{tabular}
}}
b.= baseline, c.= compressed
    \vspace{-2mm}
\end{table}

Compared to recomputation-based memory reduction solution~\cite{chen2016training, gomez2017reversible}, \techname can provide a high compression ratio to activation data of convolutional layers that cannot be reduced with acceptable overhead by recomputation.
Compared to the migration-based solution, \techname provides a higher compression ratio on our evaluated models without the constrain of CPU-GPU communication (e.g., \techname provides 11.0$\times$ compression ratio compared to 2.1$\times$ by Layrub~\cite{liu2018layrub} on ResNet-50).
\techname is also more flexible in comparison: if users decide not to increase the batch size (although it is a common optimization), \techname can still help train the same model with much lower memory requirement, which otherwise cannot be trained with the baseline training, enabling a scaled-up training solution; it can also save more precious shared-memory space for other co-running workloads via container~\cite{amaral2017topology} or GPU multi-instance technologies~\cite{MIG2020}.
Compared with the lossless-compression-based solution \cite{rhu2018compressing}, which reduces the memory usage by only 1$\sim$2$\times$, \techname outperforms it by over 9$\times$; 
compared with the current state-of-the-art JPEG-based solution \cite{evans2020jpeg}, which uses an hardware-implemented image-based compressor to provide up to 7$\times$ compression ratios, \techname outperforms it by 1.5$\times$ and 1.8$\times$ on ResNet-18 and ResNet-50, respectively, shown in Table~\ref{tab:compress}. 
Note all methods mentioned above including our proposed \techname are orthogonal to each other and can be deployed together to maximize the compression ratio and training performance.

\begin{figure}[]
    \centering
    \vspace{-2mm}
    \includegraphics[width=0.95\linewidth]{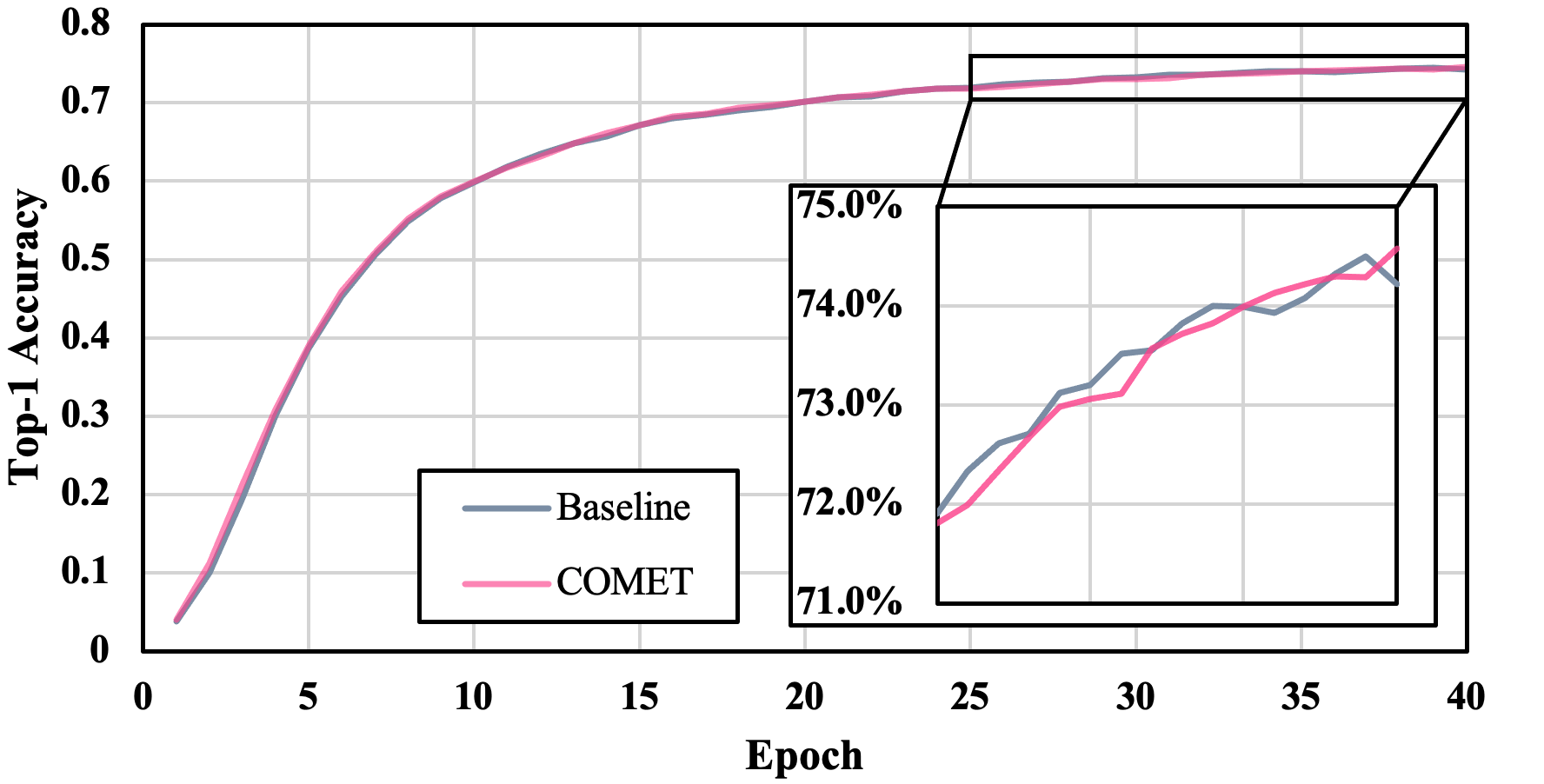}
    \vspace{-4mm}
    \caption{Comparison of validation accuracy between baseline training and \techname on EfficientNet-B0.}
    \label{fig:fig-5-efficientnet}
\end{figure}

Moreover, other than training from scratch, \techname is also capable of fine tuning from an existing model. We use EfficientNet-B0~\cite{tan2019efficientnet} for demonstration. The model was pre-trained on ImageNet dataset and is evaluated by fine tuning on the Stanford Dogs dataset that contains 12,000 images for training and 8,580 images for testing in 120 categories. We set all layers ``trainable'' and use a relatively small learning rate of $2^{-5}$ to perform \techname on all activation data. Note that compressing all convolutional layers (i.e., introducing compression error to all convolutional layers) during the fine-tuning stage with a small learning rate is more challenging to \techname compared to the partial-layer fine-tuning approach.
Similar to Figure~\ref{fig:fig-5-alexnet}, Figure~\ref{fig:fig-5-efficientnet} shows that the validation accuracy curve of the baseline and \techname are aligned with each other, meaning that \techname successfully provides a high compression ratio with minimal validation accuracy loss during the training process.

\subsection{Performance Evaluation and Analysis}

\begin{figure}[]
    \centering
    \includegraphics[width=1.0\linewidth]{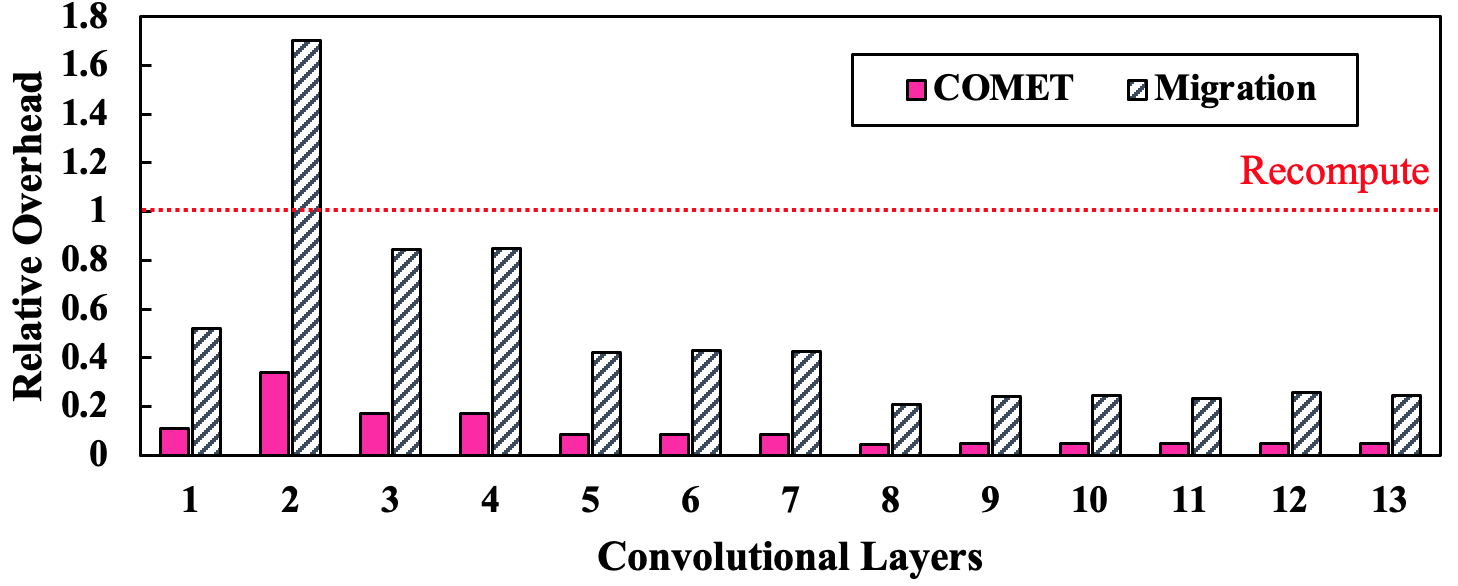}
    \vspace{-4mm}
    \caption{\textcolor{black}{Overhead comparison between migration, recomputation, and \techname on VGG-16 (batch size = 128). Time is normalized to the computation time of given convolutional layer, which is also the recompute overhead.}}
    \label{fig:fig-5-mig}
    \vspace{-2mm}
\end{figure}

\begin{figure}[]
    \centering
    \includegraphics[width=0.95\linewidth]{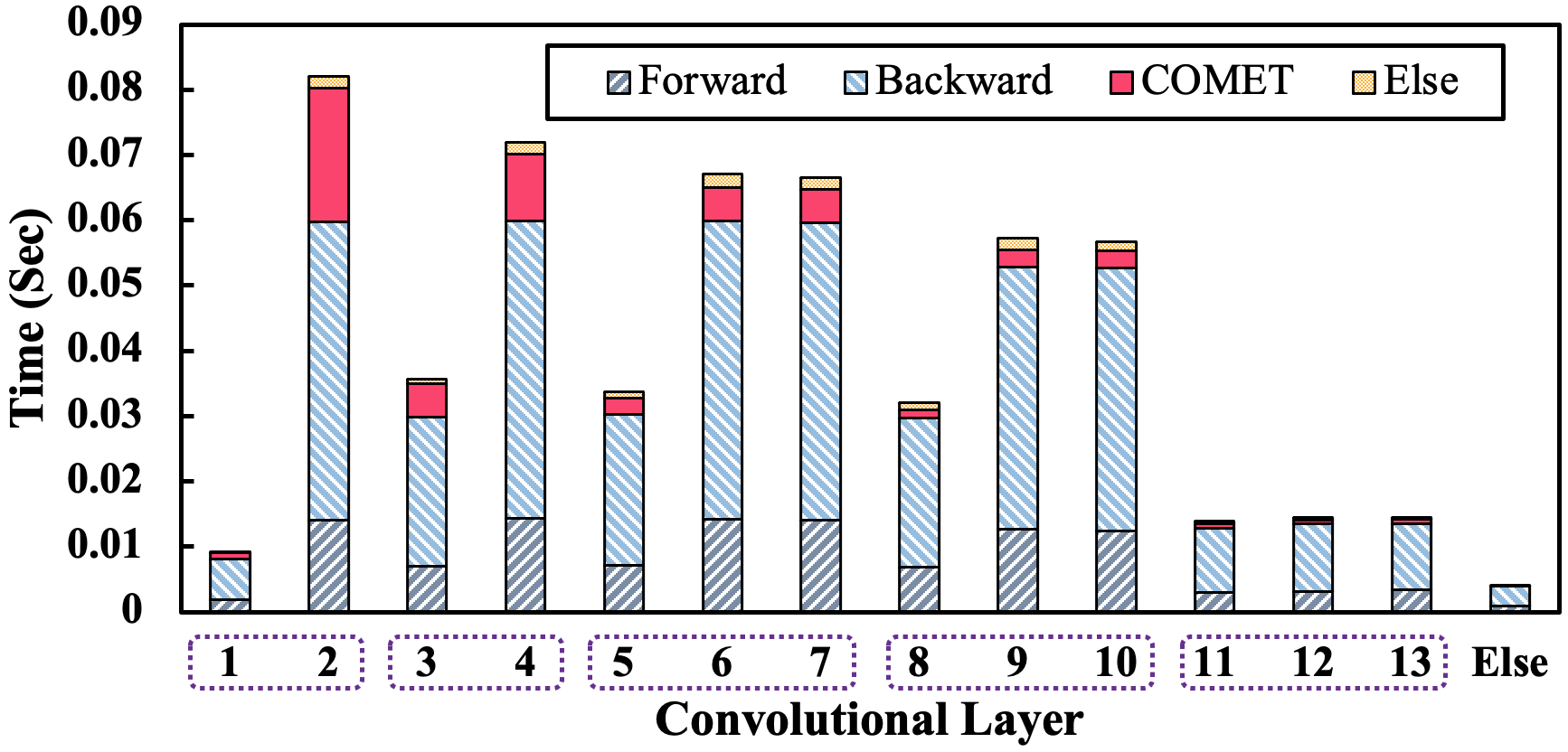}
    \vspace{-4mm}
    \caption{Time breakdown of one training iteration with \techname on VGG-16 (batch size = 128).}
    \label{fig:fig-5-breakdown}
\end{figure}

As aforementioned, \techname features a theoretical analysis to estimate error propagation and to provide an adaptive configuration for activation data compression based on easy-to-collect parameters. Note that it is almost impossible to find adaptive configurations by trial-and-error method for DNN training because of its extremely long training period, not to mention it requires a large number of \textcolor{black}{traverses} for high configuration precision. As a result, only by using the theoretical analysis as the backbone of \techname can we avoid the trial-and-error method to select adaptive solutions.

Regarding the performance overhead of \techname, it needs to extract usable parameters and compute the compression configuration every 1,000 iterations, while the amortized overhead is almost negligible;
on the other hand, thanks to its high working efficiency, cuSZ can provide an extremely high compression and decompression speed on GPU \cite{tian2020cusz}. 
We modify cuSZ by adding a filter that changes all the values under the error bound to zeros, as discussed in Section \ref{sub:comp}.
This helps us to keep zeros in the activation data unchanged while only causing little overhead to the framework. 
Overall, \techname introduces about 17\% overhead the training process while keeping the same training batch size for our experimented models. 
In comparison, the state-of-the-art recomputation-based solution~\cite{gomez2017reversible} cannot support convolutional layers without the significantly large overhead, and the image-based compression solution~\cite{evans2020jpeg} is based on hardware implementation and simulation. Moreover, the state-of-the-art migration solution Layrub achieves a memory reduction of 2.4$\times$ on average but with a higher performance overhead of 24.1\% \cite{liu2018layrub} on \textcolor{black}{K40M GPU}. 
\textcolor{black}{Figure~\ref{fig:fig-5-mig} shows the overhead comparison between the three techniques on convolutional layers. Note that the migration takes an even larger overhead than claimed because the computational power improvement from K40M to V100 reduces the computation time while both systems still bottleneck by similar communication bandwidth.}
\textcolor{black}{Overall, compared to data migration, \techname can provide a comparable compression ratio while introducing fairly less overhead;
both outperform recomputation on convolutional layers.}

Figure~\ref{fig:fig-5-breakdown} demonstrates the training time breakdown of \techname on VGG-16. We can observe that for most layers, \techname only introduces a little overhead in comparison to the forward and backward propagation time. \textcolor{black}{Specifically}, \techname introduces an overall overhead of 11.5\% with the same batch size. Note that for each group of convolutional layers (framed in purple dashed line in Figure~\ref{fig:fig-5-breakdown}), the first convolutional layer (i.e., layer 1, 3, 5, 8, and 11) has less overhead compared to the other layers in the group due to its smaller size of activation data. 
We also acknowledge that \techname may introduce higher overheads than expected to the networks that contain convolutional layers with many 1$\times$1 kernels. This is because 1$\times$1 kernels take little time to compute but require a relatively high overhead to compress and decompress. 
Calculating such layers is very efficient, compared with the GPU (de)compression on similar sizes of activation data. 
Thus, \techname is more suitable for the CNNs composed of larger convolution kernels than 1$\times$1.

\begin{figure}[]
    \centering
    \includegraphics[width=0.95\linewidth]{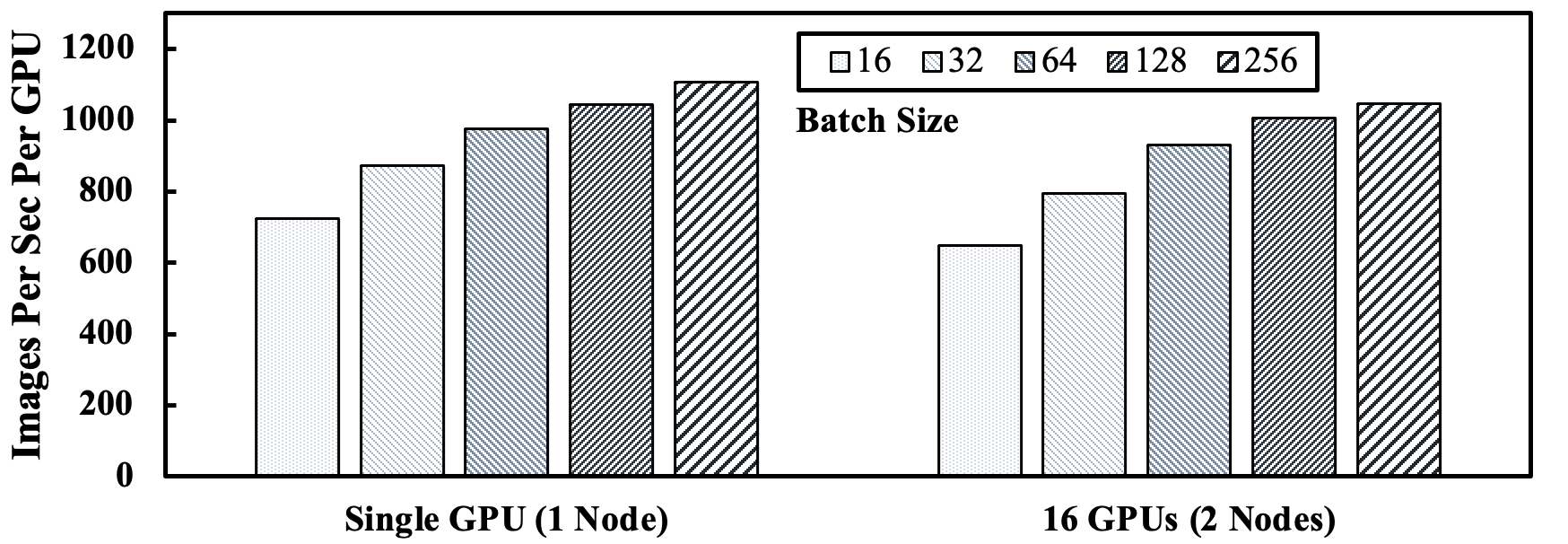}
    \vspace{-3mm}
    \caption{Training throughputs with different batch sizes on ResNet-50. Throughput per GPU slightly decreases with more GPUs due to increasing communication overhead for gradient updates.}
    \label{fig:fig-5-performance}
\end{figure}

\techname introduces relatively small overhead to the training process while can greatly \textcolor{black}{reduce} the memory utilization and allow larger and wider neural networks to be trained with limited GPU memory. Moreover, the saved memory can also be further utilized for a larger batch size, which improves the overall performance. Figure~\ref{fig:fig-5-performance} shows the improvement of training throughout (i.e., images per second per GPU) with increasing batch size on both single-/multi-GPU cases.
Specifically, \techname provides a training throughout improvement of up to 1.27$\times$ and 1.30$\times$ on ResNet-50 with 1 GPU and 8 GPUs, respectively. 
Furthermore, this performance improvement can offset the overheads of \techname (e.g., error-bound estimation and activation data compression). 
For example, the overall overhead can be reduced from 11.5\% to -7\% on VGG-16 by utilizing the saved memory to increase the batch size from 32 to 256.
\textcolor{black}{It means that we can still improve the overall training performance by fully utilizing the GPU computational performance despite the compression overhead.}

\begin{figure}[]
    \centering
    \includegraphics[width=0.9\linewidth]{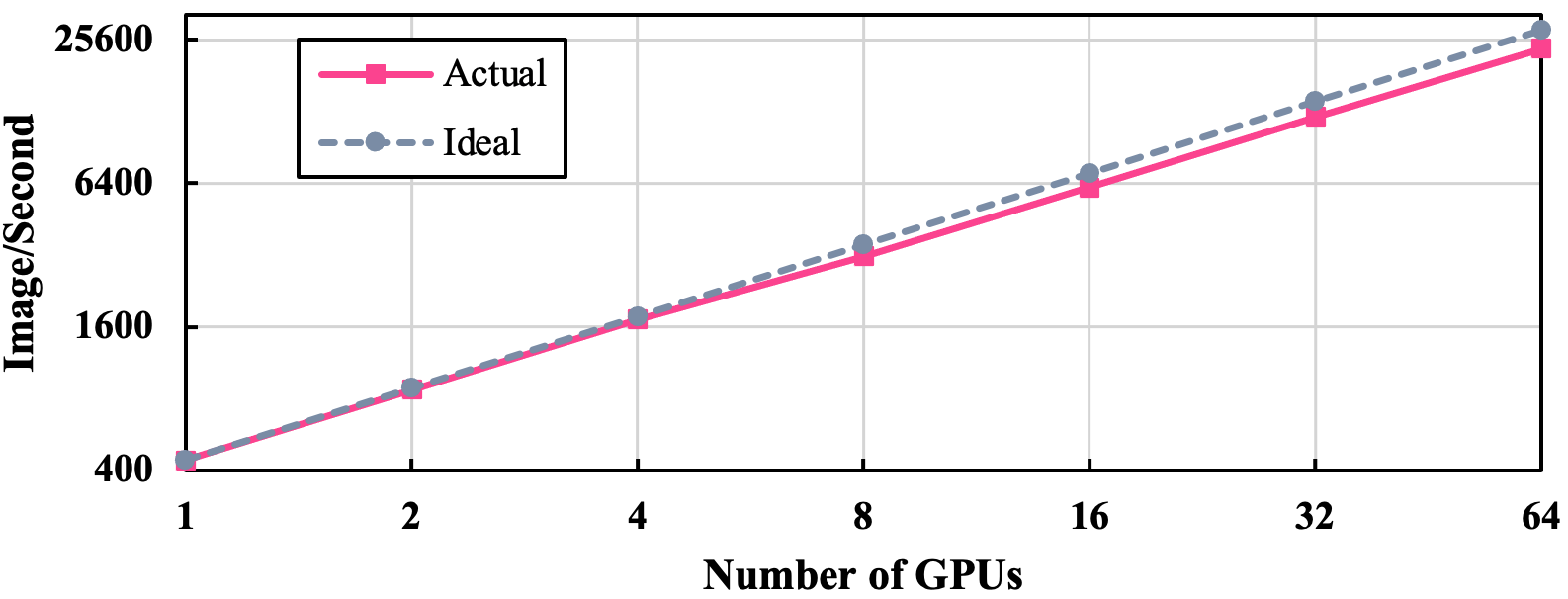}
    \vspace{-4mm}
    \caption{Multi-GPU performance of \techname framework with ResNet-50. Y-axis represents the overall throughput. ``Ideal'' represents the ideally linear performance scalability.}
    \label{fig:fig-5-scale}
\end{figure}

In addition, we also evaluate the performance of our framework scaled from a single GPU (from a single node) to 64 GPUs (from 8 nodes), as shown in Figure~\ref{fig:fig-5-scale}. It illustrates that \techname is highly scalable thanks to no extra communication cost introduced. Note that the small performance degradation compared to the ideal speedup is due to the training framework itself (e.g., the communication overhead of AllReduce for gradient update).

\begin{figure}[]
    \centering
    \includegraphics[width=1.0\linewidth]{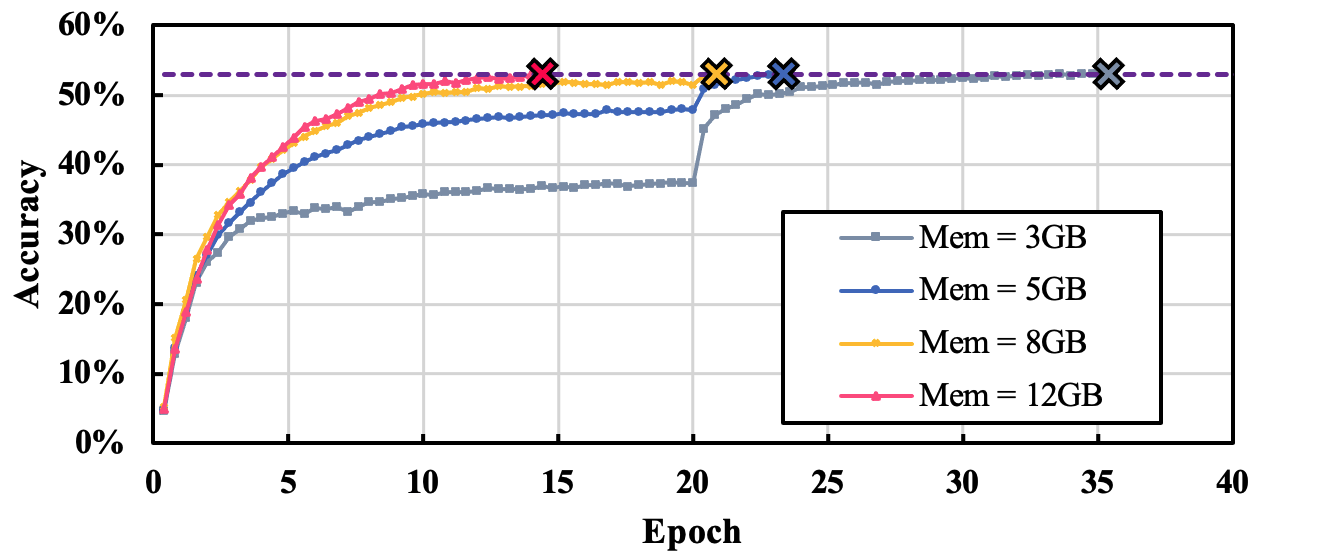}
    \vspace{-8mm}
    \caption{\textcolor{black}{Validation accuracy curve of \techname under different GPU memory constraint on AlexNet. By compressing activation data and increasing batch size, \techname can improve the end-to-end training performance.}}
    \label{fig:fig-5-batch}
\end{figure}

Finally, \techname improves the end-to-end training performance by faster convergence speed due to larger batch size~\cite{you2017scaling}. 
This is because a larger batch size involved more training data per iteration which leads to a more precise gradient direction. 
Figure~\ref{fig:fig-5-batch} shows that the convergence speed of \techname is faster with a larger batch size on AlexNet.
\textcolor{black}{For example, it takes 14.6 epochs with the batch size of 1024 (\techname at Mem = 12 GB) while 21.3 epochs with the batch size of 512 (\techname at Mem = 8 GB) to train AlexNet to the same top-1 accuracy of 53.0\%. The baseline training under 12 GB memory can only use the batch size of 512, which is significantly slower than \techname.}
As a result, we can achieve over $2\times$ speedup by using $8\times$ larger batch size for AlexNet.
\textcolor{black}{Note that the convergence speeds of AlexNet at the batch sizes of 512 and 1024 are similar to each other, as both of them reach the scalability limit, meaning that larger batch sizes cannot further improve the training performance.}

\textcolor{black}{It is worth noting that prior studies~\cite{goyal2017accurate, shallue2018measuring} showed that the scaling limit is considerably large for many deep neural networks. 
For example, Goyal \textit{et al.} work \cite{goyal2017accurate} provides tuning insights to train ResNet-50 at an enormous batch size of 8k without scaling bottleneck.
Shallue \textit{et al.}~\cite{shallue2018measuring} points out that the scaling bottleneck can be more significant for deep convolutional models.
Thus, increasing batch size can reduce the overall training time with the same amount of compute resources and significantly increase the training performance on various models and datasets~\cite{liu2018layrub,evans2020jpeg,smith2019super,pauloski2020convolutional}.
Considering that the memory consumption is relatively large for datasets like ImageNet, \techname can help the training reach the maximum batch size that can achieve the minimum time-to-solution.
}

\textcolor{black}{Overall, we identify that the performance improvement of \techname is threefold: (1) higher throughput per GPU unit due to higher resource utilization (thanks to larger batch size),
(2) higher convergence speed (thanks to larger batch size within the scaling limit),
and (3) 
a new capability of training larger models with limited memory space.
We acknowledge that not all three benefits can be achieved at the same time for a given model. For simple models such as AlexNet, enlarging the batch size to increase the convergence speed only happens at a relatively small batch size (i.e, $2^{10}$). 
}

\section{Conclusion and Future Work}
\label{sec:conclusion}

In this paper, we propose a novel memory-efficient deep learning training framework. 
We utilize the SZ error-bounded lossy compressor to reduce the memory consumption of convolutional layers. 
We develop an error propagation model and prove its accuracy. 
We evaluate our proposed framework on several popular CNNs with the ImageNet dataset.
The result shows that our framework significantly reduces the memory usage by up to 13.5$\times$ with little or no accuracy loss. 
Compared with the state-of-the-art compression-based approach, our framework can provide a memory reduction improvement of up to 1.8$\times$. 
By leveraging the saved memory, \techname can improve the end-to-end training performance (e.g., about 2$\times$ on AlexNet).
We plan to integrate 
data migration and recomputation methods into \techname for higher performance and more memory reduction.
We will also explore the applicability of \techname to other types of layers and models such as transformer.
Moreover, we will further reduce the (de)compression overhead of \techname by overlapping compression with operations such as convolution.
\section*{Acknowledgments}
This material is based upon work supported by National Science Foundation under Grant No. OAC-2034169 and OAC-2042084. The work was also partially supported by Australian Research Council Discovery Project DP210101984 and Facebook Faculty Award. 
This work used the Bridges-2 system at the Pittsburgh Supercomputing Center (PSC) under the Extreme Science and Engineering Discovery Environment (XSEDE).
The authors acknowledge the Texas Advanced Computing Center (TACC) at The University of Texas at Austin for providing access to the Longhorn system that has contributed to the research results reported within this paper.

\newpage
\bibliographystyle{ACM-Reference-Format}
\bibliography{refs}

\end{document}